# On Prediction Using Variable Order Markov Models

**Ron Begleiter**                                    RONBEG@CS.TECHNION.AC.IL

**Ran El-Yaniv**                                      RANI@CS.TECHNION.AC.IL
*Department of Computer Science*
*Technion - Israel Institute of Technology*
*Haifa 32000, Israel*

**Golan Yona**                                        GOLAN@CS.CORNELL.EDU
*Department of Computer Science*
*Cornell University*
*Ithaca, NY 14853, USA*

## Abstract

This paper is concerned with algorithms for prediction of discrete sequences over a finite alphabet, using variable order Markov models. The class of such algorithms is large and in principle includes any lossless compression algorithm. We focus on six prominent prediction algorithms, including Context Tree Weighting (CTW), Prediction by Partial Match (PPM) and Probabilistic Suffix Trees (PSTs). We discuss the properties of these algorithms and compare their performance using real life sequences from three domains: proteins, English text and music pieces. The comparison is made with respect to prediction quality as measured by the average log-loss. We also compare classification algorithms based on these predictors with respect to a number of large protein classification tasks. Our results indicate that a "decomposed" CTW (a variant of the CTW algorithm) and PPM outperform all other algorithms in sequence prediction tasks. Somewhat surprisingly, a different algorithm, which is a modification of the Lempel-Ziv compression algorithm, significantly outperforms all algorithms on the protein classification problems.

## 1. Introduction

Learning of sequential data continues to be a fundamental task and a challenge in pattern recognition and machine learning. Applications involving sequential data may require prediction of new events, generation of new sequences, or decision making such as classification of sequences or sub-sequences. The classic application of discrete sequence prediction algorithms is *lossless compression* (Bell et al., 1990), but there are numerous other applications involving sequential data, which can be directly solved based on effective prediction of discrete sequences. Examples of such applications are biological sequence analysis (Bejerano & Yona, 2001), speech and language modeling (Schütze & Singer, 1994; Rabiner & Juang, 1993), text analysis and extraction (McCallum et al., 2000) music generation and classification (Pachet, 2002; Dubnov et al., 2003), hand-writing recognition (Singer & Tishby, 1994) and behaviormetric identification (Clarkson & Pentland, 2001; Nisenson et al., 2003).[1]

---

1. Besides such applications, which directly concern sequences, there are other, less apparent applications, such as image and texture analysis and synthesis that can be solved based on discrete sequence prediction algorithms; see, e.g., (Bar-Joseph et al., 2001).

The literature on prediction algorithms for discrete sequences is rather extensive and offers various approaches to analyze and predict sequences over finite alphabets. Perhaps the most commonly used techniques are based on Hidden Markov Models (HMMs) (Rabiner, 1989). HMMs provide flexible structures that can model complex sources of sequential data. However, dealing with HMMs typically requires considerable understanding of and insight into the problem domain in order to restrict possible model architectures. Also, due to their flexibility, successful training of HMMs usually requires very large training samples (see hardness results of Abe & Warmuth, 1992, for learning HMMs).

In this paper we focus on general-purpose prediction algorithms, based on learning *Variable order Markov Models (VMMs)* over a finite alphabet $\Sigma$. Such algorithms attempt to learn probabilistic finite state automata, which can model sequential data of considerable complexity. In contrast to $N$-gram Markov models, which attempt to estimate conditional distributions of the form $P(\sigma|s)$, with $s \in \Sigma^N$ and $\sigma \in \Sigma$, VMM algorithms learn such conditional distributions where context lengths $|s|$ vary in response to the available statistics in the training data. Thus, VMMs provide the means for capturing both large and small order Markov dependencies based on the observed data. Although in general less expressive than HMMs, VMM algorithms have been used to solve many applications with notable success. The simpler nature of VMM methods also makes them amenable for analysis, and some VMM algorithms that we discuss below enjoy tight theoretical performance guarantees, which in general are not possible in learning using HMMs.

There is an intimate relation between prediction of discrete sequences and lossless compression algorithms, where, in principle, any lossless compression algorithm can be used for prediction and vice versa (see, e.g., Feder & Merhav, 1994). Therefore, there exist an abundance of options when choosing a prediction algorithm. This multitude of possibilities poses a problem for practitioners seeking a prediction algorithm for the problem at hand.

Our goal in this paper is to consider a number of VMM methods and compare their performance with respect to a variety of practical prediction tasks. We aim to provide useful insights into the choice of a good general-purpose prediction algorithm. To this end we selected six prominent VMM algorithms and considered sequential data from three different domains: molecular biology, text and music. We focus on a setting in which a learner has the start of some sequence (e.g., a musical piece) and the goal is to generate a model capable of predicting the rest of the sequence. We also consider a scenario where the learner has a set of training samples from some domain and the goal is to predict, as accurately as possible, new sequences from the same domain. We measure performance using log-loss (see below). This loss function, which is tightly related to compression, measures the quality of probabilistic predictions, which are required in many applications.

In addition to these prediction experiments we examine the six VMM algorithms with respect to a number of large protein classification problems. Protein classification is one of the most important problems in computational molecular biology. Such classification is useful for functional and structural categorization of proteins, and may direct experiments to further explore and study the functionality of proteins in the cell and their interactions with other molecules.

In the prediction experiments, two of the VMM algorithms we consider consistently achieve the best performance. These are the 'decomposed context tree weighting (CTW)' and 'prediction by partial match (PPM)' algorithms. CTW and PPM are well known in the lossless compression arena as outstanding players. Our results thus provide further evidence of the superiority of these algorithms, with respect to new domains and two different prediction settings. The results of our protein classification experiments are rather surprising as both of these two excellent predictors are inferior to an algorithm, which is obtained by simple modifications of the prediction component of the well-known Lempel-Ziv-78 compression algorithm (Ziv & Lempel, 1978; Langdon, 1983). This rather new algorithm, recently proposed by Nisenson et al. (2003), is a consistent winner in the all the protein classification experiments and achieves surprisingly good results that may be of independent interest in protein analysis.

This paper is organized as follows. We begin in Section 2 with some preliminary definitions. In Section 3, we present six VMM prediction algorithms. In Section 4, we present the experimental setup. In Sections 5-6, we discuss the experimental results. Part of the experimental related details are discussed in Appendices A-C. In Section 7, we discuss the related work. In Section 8, we conclude by introducing a number of open problems raised by this research. Note that the source code of all the algorithms we consider is available at `http://www.cs.technion.ac.il/~ronbeg/vmm`.

## 2. Preliminaries

Let $\Sigma$ be a finite alphabet. A learner is given a training sequence $q_1^n = q_1 q_2 \cdots q_n$, where $q_i \in \Sigma$ and $q_i q_{i+1}$ is the concatenation of $q_i$ and $q_{i+1}$. Based on $q_1^n$, the goal is to learn a model $\hat{P}$ that provides a probability assignment for any future outcome given some past. Specifically, for any "context" $s \in \Sigma^*$ and symbol $\sigma \in \Sigma$ the learner should generate a conditional probability distribution $\hat{P}(\sigma|s)$.

Prediction performance is measured via the *average log-loss* $\ell(\hat{P}, x_1^T)$ of $\hat{P}(\cdot|\cdot)$, with respect to a test sequence $x_1^T = x_1 \cdots x_T$,

$$\ell(\hat{P}, x_1^T) = -\frac{1}{T} \sum_{i=1}^{T} \log \hat{P}(x_i|x_1 \cdots x_{i-1}), \tag{1}$$

where the logarithm is taken to base 2. Clearly, the average log-loss is directly related to the likelihood $\hat{P}(x_1^T) = \prod_{i=1}^{T} \hat{P}(x_i|x_1 \cdots x_{i-1})$ and minimizing the average log-loss is completely equivalent to maximizing a probability assignment for the entire test sequence.

Note that the converse is also true. A *consistent* probability assignment $\hat{P}(x_1^T)$, for the entire sequence, which satisfies $\hat{P}(x_1^{t-1}) = \sum_{x_t \in \Sigma} \hat{P}(x_1 \cdots x_{t-1} x_t)$, for all $t = 1, \ldots, T$, induces conditional probability assignments,

$$\hat{P}(x_t|x_1^{t-1}) = \hat{P}(x_1^t)/\hat{P}(x_1^{t-1}), \quad t = 1, \ldots, T.$$

Therefore, in the rest of the paper we interchangeably consider $\hat{P}$ as a conditional distribution or as a complete (consistent) distribution of the test sequence $x_1^T$.

The log-loss has various meanings. Perhaps the most important one is found in its equivalence to lossless compression. The quantity $-\log \hat{P}(x_i|x_1 \cdots x_{i-1})$, which is also called the 'self-information', is the ideal compression or "code length" of $x_i$, in bits per symbol, with respect to the conditional distribution $\hat{P}(X|x_1 \cdots x_{i-1})$, and can be implemented online (with arbitrarily small redundancy) using arithmetic encoding (Rissanen & Langdon, 1979).

Thus, the average log-loss also measures the average compression rate of the test sequence, when using the predictions generated by $\hat{P}$. In other words, a small average log-loss over the $x_1^T$ sequence implies a good compression of this sequence.[2]

Within a probabilistic setting, the log-loss has the following meaning. Assume that the training and test sequences were emitted from some unknown probabilistic source $P$. Let the test sequence be given by the value of the sequence of random variables $X_1^T = X_1 \cdots X_T$. A well known fact is that the distribution $P$ uniquely minimizes the mean log-loss; that is,[3]

$$P = \arg\min_{\hat{P}} \left\{ -\mathbf{E}_P \{ \log \hat{P}(X_1^T) \} \right\}.$$

Due to the equivalence of the log-loss and compression, as discussed above, the mean of the log-loss of $P$ (under the true distribution $P$) achieves the best possible compression, or the entropy $H_T(P) = -\mathbf{E} \log P(X_1^T)$. However, the true distribution $P$ is unknown and the learner generates the proxy $\hat{P}$ using the training sequence. The extra loss (due to the use of $\hat{P}$, instead of $P$) beyond the entropy, is called the *redundancy* and is given by

$$D_T(P||\hat{P}) = \mathbf{E}_P \left\{ -\log \hat{P}(X_1^T) - (-\log P(X_1^T)) \right\}. \tag{2}$$

It is easy to see that $D_T(P||\hat{P})$ is the ($T^{th}$ order) Kullback-Leibler (KL) divergence (see Cover & Thomas, 1991, Sec. 2.3). The *normalized redundancy* $D_T(P||\hat{P})/T$ (of a sequence of length $T$) gives the extra bits per symbol (over the entropy rate) when compressing a sequence using $\hat{P}$.

This probabilistic setting motivates a desirable goal, when devising a general-purpose prediction algorithm: minimize the redundancy uniformly, with respect to all possible distributions. A prediction algorithm for which we can bound the redundancy uniformly, with respect to all distributions in some given class, is often called *universal* with respect to that class. A lower bound on the redundancy of any universal prediction (and compression) algorithm is $\Omega(K(\frac{\log T}{2T}))$, where $K$ is (roughly) the number of parameters of the model encoding the distribution $\hat{P}$ (Rissanen, 1984).[4] Some of the algorithms we discuss below are universal with respect to some classes of sources. For example, when an upper bound on the Markov order is known, the CTW algorithm (see below) is universal (with respect to ergodic and stationary sources), and in fact, has bounded redundancy, which is close to the Rissannen lower bound.

## 3. VMM Prediction Algorithms

In order to assess the significance of our results it is important to understand the sometimes subtle differences between the different algorithms tested in this study and their variations. In this section we describe in detail each one of these six algorithms. We have adhered to a unified notation schema in an effort to make the commonalities and differences between these algorithms clear.

---

2. The converse is also true: any compression algorithm can be translated into probability assignment $\hat{P}$; see, e.g., (Rissanen, 1984).

3. Note that very specific loss functions satisfy this property. (See Miller et al., 1993).

4. A related lower bound in terms of channel capacity is given by Merhav and Feder (1995).

Most algorithms for learning VMMs include three components: counting, smoothing (probabilities of unobserved events) and variable-length modeling. Specifically, all such algorithms base their probability estimates on counts of the number of occurrences of symbols $\sigma$ appearing after contexts $s$ in the training sequence. These counts provide the basis for generating the predictor $\hat{P}$. The smoothing component defines how to handle unobserved events (with zero value counts). The existence of such events is also called the "zero frequency problem". Not handling the zero frequency problem is harmful because the log-loss of an unobserved but possible event, which is assigned a zero probability by $\hat{P}$, is infinite. The algorithms we consider handle such events with various techniques.

Finally, variable length modeling can be done in many ways. Some of the algorithms discussed here construct a single model and some construct several models and average them. The models themselves can be bounded by a pre-determined constant bound, which means that the algorithm does not consider contexts that are longer than the bound. Alternatively, models may not be bounded a-priori, in which case the maximal context size is data-driven.

There are a great many VMM prediction algorithms. In fact, any lossless compression algorithm can be used for prediction. For the present study we selected six VMM prediction algorithms, described below. We attempted to include algorithms that are considered to be top performers in lossless compression. We, therefore, included the 'context tree weighting (CTW)' and 'prediction by partial match (PPM)' algorithms. The 'probabilistic suffix tree (PST)' algorithm is well known in the machine learning community. It was successfully used in a variety of applications and is hence included in our set. To gain some perspective we also included the well known LZ78 (prediction) algorithm that forms the basis of many commercial applications for compression. We also included a recent prediction algorithm from Nisenson et al. (2003) that is a modification of the LZ78 prediction algorithm. The algorithms we selected are quite different in terms of their implementations of the three components described above. In this sense, they represent different approaches for VMM prediction.[5]

## 3.1 Lempel-Ziv 78 (LZ78)

The LZ78 algorithm is among the most popular lossless compression algorithms (Ziv & Lempel, 1978). It is used as the basis of the Unix `compress` utility and other popular archiving utilities for PCs. It also has performance guarantees within several analysis models. This algorithm (together with the LZ77 compresion method) attracted enormous attention and inspired the area of lossless compression and sequence prediction.

The *prediction* component of this algorithm was first discussed by Langdon (1983) and Rissanen (1983). The presentation of this algorithm is simplified after the well-known LZ78 *compression* algorithm, which works as follows, is understood. Given a sequence $q_1^n \in \Sigma^n$, LZ78 incrementally parses $q_1^n$ into non-overlapping adjacent 'phrases', which are collected into a phrase 'dictionary'. The algorithm starts with a dictionary containing the empty phrase $\epsilon$. At each step the algorithm parses a new phrase, which is the shortest phrase that is not yet in the dictionary. Clearly, the newly parsed phrase $s'$ extends an existing dictionary phrase by one symbol; that is, $s' = s\sigma$, where $s$ is already in the dictionary ($s$ can be the empty phrase). For compression, the algorithm encodes the index of $s'$ (among all parsed phrases) followed by a fixed code for $\sigma$. Note that coding issues will not concern us in this paper. Also observe that LZ78 compresses sequences without explicit probabilistic estimates. Here is an example of this LZ78 parsing: if $q_1^{11} = $ `abracadabra`, then the parsed phrases are `a|b|r|ac|ad|ab|ra`. Observe that the empty sequence $\epsilon$ is always in the dictionary and is omitted in our discussions.

---

5. We did not include in the present work the prediction algorithm that can be derived from the more recent `bzip` compression algorithm (see `http://www.digistar.com/bzip2`), which is based on the successful Burrows-Wheeler Transform (Burrows & Wheeler, 1994; Manzini, 2001).

An LZ78-based prediction algorithm was proposed by Langdon (1983) and Rissanen (1983). We describe separately the learning and prediction phases.[6] For simplicity we first discuss the binary case where $\Sigma = \{0, 1\}$, but the algorithm can be naturally extended to alphabets of any size (and in the experiments discussed below we do use the multi-alphabet algorithm). In the learning phase the algorithm constructs from the training sequence $q_1^n$ a binary tree (trie) that records the parsed phrases (as discussed above). In the tree we also maintain counters that hold statistics of $q_1^n$. The initial tree contains a root and two (left and right) leaves. The left child of a node corresponds to a parsing of '0' and the right child corresponds to a parsing of '1'. Each node maintains a counter. The counter in a leaf is always set to 1. The counter in an internal node is always maintained so that it equals the sum of its left and right child counters. Given a newly parsed phrase $s'$, we start at the root and traverse the tree according to $s'$ (clearly the tree contains a corresponding path, which ends at a leaf). When reaching a leaf, the tree is expanded by making this leaf an internal node and adding two leaf-sons to this new internal node. The counters along the path to the root are updated accordingly.

To compute the estimate $\hat{P}(\sigma|s)$ we start from the root and traverse the tree according to $s$. If we reach a leaf before "consuming" $s$ we continue this traversal from the root, etc. Upon completion of this traversal (at some internal node, or a leaf) the prediction for $\sigma = $ '0' is the '0' (left) counter divided by the sum of '0' and '1' counters at that node, etc.

For larger alphabets, the algorithm is naturally extended such that the phrases are stored in a multi-way tree and each internal node has exactly $k = |\Sigma|$ children. In addition, each node has $k$ counters, one for each possible symbol. In Figure 1 we depict the resulting tree for the training sequence $q_1^{11} = $ `abracadabra` and calculate the probability $\hat{P}($`b`$|$`ab`$)$, assuming $\Sigma = \{a, b, c, d, r\}$.

Several performance guarantees were proven for the LZ78 compression (and prediction) algorithm. Within a probabilistic setting (see Section 2), when the unknown source is stationary and ergodic Markov of finite order, the redundancy is bounded above by $(1/\ln n)$ where $n$ is the length of the training sequence (Savari, 1997). Thus, the LZ78 algorithm is a universal prediction algorithm with respect to the large class of stationary and ergodic Markov sources of finite order.

## 3.2 Prediction by Partial Match (PPM)

The Prediction by Partial Match (PPM) algorithm (Cleary & Witten, 1984) is considered to be one of the best lossless compression algorithms.[7] The algorithm requires an upper bound $D$ on the maximal Markov order of the VMM it constructs. PPM handles the zero frequency problem using two mechanisms called *escape* and *exclusion*. There are several PPM variants distinguished by the implementation of the escape mechanism. In all variants the escape mechanism works as follows. For each context $s$ of length $k \leq D$, we allocate a probability mass $\hat{P}_k(escape|s)$ for all symbols that did not appear after the context $s$ (in the training sequence). The remaining mass $1 - \hat{P}_k(escape|s)$ is distributed among all other symbols that have non-zero counts for this context. The particular mass allocation for 'escape' and the particular mass distribution $\hat{P}_k(\sigma|s)$, over these other symbols $\sigma$, determine the PPM variant. The mechanism of all PPM variants satisfies the following (recursive) relation,

$$\hat{P}(\sigma|s_{n-D+1}^n) = \begin{cases} \hat{P}_D(\sigma|s_{n-D+1}^n), & \text{if } s_{n-D+1}^n\sigma \text{ apeared in} \\ & \text{the training sequence;} \\ \hat{P}_D(escape|s_{n-D+1}^n) \cdot \hat{P}(\sigma|s_{n-D+2}^n), & \text{otherwise.} \end{cases} \quad (3)$$

---

6. These "phases" can be combined and operated together online.

7. Specifically, the PPM-II variant of PPM currently achieves the best compression rates over the standard Calgary Corpus benchmark (Shkarin, 2002).

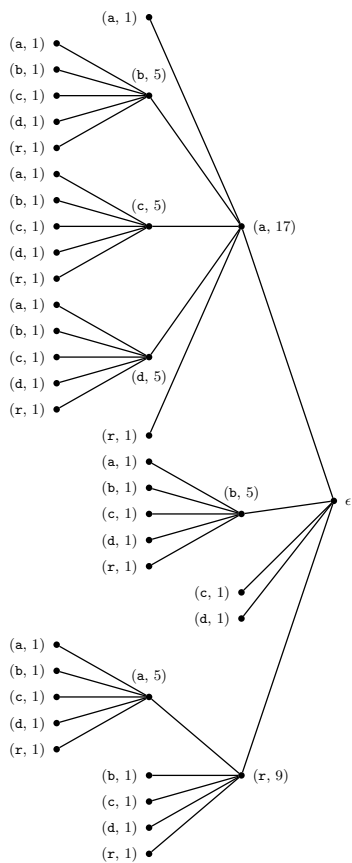

Figure 1: The tree constructed by LZ78 using the training sequence $q_1^{11} = \texttt{abracadabra}$. The algorithm extracts the set $\{\texttt{a,b,r,ac,ad,ab,ra}\}$ of "phrases" (this set is traditionally called a 'dictionary') and constructs the phrase tree accordingly. Namely, there is an internal node for every string in the dictionary, and every internal node has all $|\Sigma|$ children. For calculating $\hat{P}(\texttt{b|ar})$ we traverse the tree as follows: $\epsilon \rightarrow \texttt{a} \rightarrow \texttt{r} \rightarrow \epsilon \rightarrow \texttt{b}$ and conclude with $\hat{P}(\texttt{b|ar}) = 5/33 = 0.152$. To calculate $\hat{P}(\texttt{d|ra})$ we traverse the tree as follow: $\epsilon \rightarrow \texttt{r} \rightarrow \texttt{a} \rightarrow \texttt{d}$ and conclude with $\hat{P}(\texttt{d|ra}) = 1/5 = 0.2$.

For the empty context $\epsilon$, PPM takes $\hat{P}(\sigma|\epsilon) = 1/|\Sigma|$.[8]

The exclusion mechanism is used to enhance the escape estimation. It is based on the observation that if a symbol $\sigma$ appears after the context $s$ of length $k \leq D$, there is no need to consider $\sigma$ as part of the alphabet when calculating $\hat{P}_k(\cdot|s')$ for all $s'$ suffix of $s$ (see Equation (3)). Therefore, the estimates $\hat{P}_k(\cdot|s)$ are potentially more accurate since they are based on a smaller (observed) alphabet.

The particular PPM variant we consider here is called 'Method C' (PPM-C) and is defined as follows. For each sequence $s$ and symbol $\sigma$ let $N(s\sigma)$ denote the number of occurrences of $s\sigma$ in the training sequence. Let $\Sigma_s$ be the set of symbols appearing after the context $s$ (in the training sequence); that is, $\Sigma_s = \{\sigma \ : \ N(s\sigma) > 0\}$. For PPM-C we thus have

$$\hat{P}_k(\sigma|s) = \frac{N(s\sigma)}{|\Sigma_s| + \sum\limits_{\sigma' \in \Sigma_s} N(s\sigma')} \quad , \text{if } \sigma \in \Sigma_s; \quad (4)$$

$$\hat{P}_k(escape|s) = \frac{|\Sigma_s|}{|\Sigma_s| + \sum\limits_{\sigma' \in \Sigma_s} N(s\sigma')}, \quad (5)$$

where we assume that $|s| = k$. This implementation of the escape mechanism by Moffat (1990) is considered to be among the better PPM variants, based on empirical examination (Bunton, 1997). As noted by Witten and Bell (1991) there is no principled justification for any of the various PPM escape mechanisms.

One implementation of PPM-C is based on a trie data structure. In the learning phase the algorithm constructs a trie $\mathcal{T}$ from the training sequence $q_1^n$. Similar to the LZ78 trie, each node in $\mathcal{T}$ is associated with a symbol and has a counter. In contrast to the unbounded LZ78 trie, the maximal depth of $\mathcal{T}$ is $D+1$. The algorithm starts with a root node corresponding to the empty sequence $\epsilon$ and incrementally parses the training sequence, one symbol at a time. Each parsed symbol $q_i$ and its $D$-sized context, $x_{i-D}^{i-1}$, define a potential path in $\mathcal{T}$, which is constructed, if it does not yet exist. Note that after parsing the first $D$ symbols, each newly constructed path is of length $D+1$. The counters along this path are incremented. Therefore, the counter of any node, with a corresponding path $s\sigma$ (where $\sigma$ is the symbol associated with the node) is $N(s\sigma)$.

Upon completion, the resulting trie induces the probability $\hat{P}(\sigma|s)$ for each symbol $\sigma$ and context $s$ with $|s| \leq D$. To compute $\hat{P}(\sigma|s)$ we start from the root and traverse the tree according to the longest suffix of $s$, denoted $s'$, such that $s'\sigma$ corresponds to a complete path from the root to a leaf. We then use the counters $N(s'\sigma')$ to compute $\Sigma_{s'}$ and the estimates as given in Equations (3), (4) and (5).

The above implementation (via a trie) is natural for the learning phase and it is easy to see that the time complexity for learning $q_1^n$ is $O(n)$ and the space required for the worst case trie is $O(D \cdot n)$. However, a straightforward computation of $\hat{P}(\sigma|s)$ using the trie incurs $O(|s|^2)$ time.[9] In Figure 2 we depict the resulting tree for the training sequence $q_1^{11} = \texttt{abracadabra}$ and calculate the probabilities $\hat{P}(\texttt{b}|\texttt{ar})$ and $\hat{P}(\texttt{d}|\texttt{ra})$, assuming $\Sigma = \{\texttt{a,b,c,d,r}\}$.

---

8. It also makes sense to use the frequency count of symbols over the training sequence.

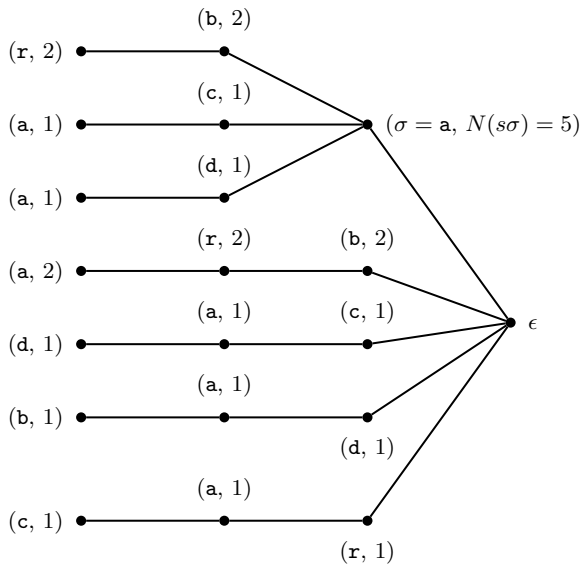

Figure 2: The tree constructed by PPM-C using the training sequence $q_1^{11} = \texttt{abracadabra}$. $\hat{P}(\texttt{d}|\texttt{ra}) = \hat{P}(escape|\texttt{ra}) \cdot \hat{P}(\texttt{d}|\texttt{a}) = \frac{1}{2} \cdot \frac{1}{4+3} = 0.07$ without using the exclusion mechanism. When using the exclusion, observe that $N(\texttt{rac}) > 0$; therefore, we can exclude $c$ when evaluating $\hat{P}(\texttt{d}|\texttt{ra}) = \hat{P}(escape|\texttt{ra}) \cdot \hat{P}(\texttt{d}|\texttt{a}) = \frac{1}{2} \cdot \frac{1}{6} = 0.08$.

### 3.3 The Context Tree Weighting Method (CTW)

The Context Tree Weighting Method (CTW) algorithm (Willems et al., 1995) is a strong lossless compression algorithm that is based on a clever idea for combining exponentially many VMMs of bounded order.[10] In this section we consider the original CTW algorithm for *binary* alphabets. In the next section we discuss extensions for larger alphabets.

If the unknown source of the given sample sequences is a "tree source" of bounded order, then the compression redundancy of CTW approaches the Rissanen lower bound at rate $1/n$.[11] A $D$-bounded *tree source* is a pair $(M, \Theta_M)$ where $M$ is set of sequences of length $\leq D$. This set must be a *complete and proper suffix set*, where 'complete' means that for any sequence $x$ of length $\geq D$, there exists $s \in M$, which is a suffix of $x$. 'Proper' means that no sequence in $M$ has a strict suffix in $M$. The set $\Theta_M$ contains one probability distribution over $\Sigma$ for each sequence in $M$. A proper and complete suffix set is called a *model*. Given some initial sequence ("state") $s \in M$, the tree source can generate a random sequence $x$ by continually drawing the next symbol from the distribution associated with the unique suffix of $x$.

---

9. After the construction of the trie it is possible to generate a corresponding suffix tree that allows a rapid computation of $\hat{P}(\sigma|s)$ in $O(|s|)$ time. It is possible to generate this suffix tree in time proportional to the sum of all paths in the trie.

10. The paper presenting this idea (Willems et al., 1995) received the 1996 Paper Award of the IEEE Information Theory Society.

11. Specifically, for $\Sigma = \{0, 1\}$, Rissanen's lower bound on the redundancy when compressing $x_1^n$ is $\Omega(|M|(\frac{\log n}{2n}))$. The guaranteed CTW redundancy is $O(|M|(\frac{\log n}{2n}) + \frac{1}{n}|M| \log |M|)$, where $M$ is the suffix set of the unknown source.

Any full binary tree of height $\leq D$ corresponds to one possible $D$-bounded model $M$. Any choice of appropriate probability distributions $\Theta_M$ defines, together with $M$, a unique tree source. Clearly, all prunings of the perfect binary tree[12] of height $D$ that are full binary trees correspond to the collection of all $D$-bounded models.

For each model $M$, CTW estimates a set of probability distributions, $\Theta_M = \{\hat{P}_M(\cdot|s)\}_{s \in M}$, such that $(M, \Theta_M)$ is a tree source. Each distribution $\hat{P}_M(\cdot|s)$ is a smoothed version of a maximum likelihood estimate based on the training sequence. The core idea of the CTW algorithm is to generate a predictor that is a mixture of all these tree sources (corresponding to all these $D$-bounded models). Let $\mathcal{M}_D$ be the collection of all $D$-bounded models. For any sequence $x_1^T$ the CTW estimated probability for $x_1^T$ is given by

$$\hat{P}_{\text{CTW}}(x_1^T) = \sum_{M \in \mathcal{M}_D} w(M) \cdot \hat{P}_M(x_1^T); \tag{6}$$

$$\hat{P}_M(x_1^T) = \prod_{i=1}^{T} \hat{P}_M(x_i|\text{suffix}_M(x_{i-D}^{i-1})), \tag{7}$$

where $\text{suffix}_M(x)$ is the (unique) suffix of $x$ in $M$ and $w(M)$ is an appropriate probability weight function. Clearly, the probability $\hat{P}_M(x_1^T)$, given in (7), is a product of independent events, by the definition of a tree source. Also note that we assume that the sequence $x_1^T$ has an "historical" context defining the conditional probabilities $\hat{P}(x_i|\text{suffix}_M(x_{i-D}^{i-1}))$ of the first symbols. For example, we can assume that the sequence $x_1^T$ is a continuation of the training sequence. Alternatively, we can ignore (up to) the first $D$ symbols of $x_1^T$. A more sophisticated solution is proposed by Willems (1998); see also Section 7.

Note that there is a huge (exponential in $D$) number of elements in the mixture (6), corresponding to all bounded models. To describe the CTW prediction algorithm we need to describe (i) how to (efficiently) estimate all the distributions $\hat{P}_M(\sigma|s)$; (ii) how to select the weights $w(M)$; and (iii) how to efficiently compute the CTW mixture (6).

For each model $M$, CTW computes the estimate $\hat{P}_M(\cdot|s)$ using the Krichevsky-Trofimov (KT) estimator for memoryless binary sources (Krichevsky & Trofimov, 1981). The KT-estimator enjoys an optimal proven redundancy bound of $\frac{2+\log n}{2n}$. This estimator, which is very similar to Laplace's law of succession, is also very easy to calculate. Given a training sequence $q \neq \epsilon$, the KT-estimator, based on $q$, is equivalent to

$$\hat{P}_{\text{KT}}(0|q) = \frac{N_0(q) + \frac{1}{2}}{N_0(q) + N_1(q) + 1};$$

$$\hat{P}_{\text{KT}}(1|q) = 1 - \hat{P}_{\text{KT}}(0|q),$$

---

12. In a *perfect binary tree* all internal nodes have two sons and all leaves are at the same level.

where $N_0(q) = N(0)$ and $N_1(q) = N(1)$. That is, $\hat{P}_{\text{KT}}(0|q)$ is the KT-estimated probability for '0', based on a frequency count in $q$.[13]

For any sequence $s$ define $\text{sub}_s(q)$ to be the unique (non-contiguous) sub-sequence of $q$ consisting of the symbols following the occurrences of $s$ in $q$. Namely, it is the concatenation of all symbols $\sigma$ (in the order of their occurrence in $q$) such that $s\sigma$ is a substring of $q$. For example, if $s = 101$ and $q = 101011010$, then $\text{sub}_{101}(q) = 010$. Using this definition we can calculate the KT-estimated probability of a test sequence $x_1^T$ according to a context $s$, based on the training sequence $q$. Letting $q_s = \text{sub}_s(q)$ we have,

$$\hat{P}_{\text{KT}}^s(x_1^T|q) = \prod_{i=1}^{T} \hat{P}_{\text{KT}}(x_i|q_s).$$

For the empty sequence $\epsilon$ we have $\hat{P}_{\text{KT}}^s(\epsilon|q) = 1$.

One of the main technical ideas in the CTW algorithm is an efficient recursive computation of Equation (6). This is achieved through the following recursive definition. For any $s$ such that $|s| \leq D$, define

$$\hat{P}_{\text{CTW}}^s(x_1^T) = \begin{cases} \frac{1}{2}\hat{P}_{\text{KT}}^s(x_1^T|q) + \frac{1}{2}\hat{P}_{\text{CTW}}^{0s}(x_1^T)\hat{P}_{\text{CTW}}^{1s}(x_1^T), & \text{if } |s| < D; \\ \hat{P}_{\text{KT}}^s(x_1^T|q), & \text{otherwise } (|s| = D), \end{cases} \qquad (8)$$

where $q$ is, as usual, the training sequence. As shown by Willems et al. (1995, Lemma 2), for any training sequence $q$ and test sequence $x$, $\hat{P}_{\text{CTW}}^\epsilon(x)$ is precisely the CTW mixture as given in Equation (6), where the probability weights $w(M)$ are[14]

$$w(M) = 2^{-C_D(M)};$$
$$C_D(M) = |\{s \in M\}| - 1 + |\{s \in M : |s| < D\}|.$$

In the learning phase, the CTW algorithm processes a training sequence $q_1^n$ and constructs a binary context tree $\mathcal{T}$. An outgoing edge to a left son in $\mathcal{T}$ is labelled by '0' and an outgoing edge to a right son, by '1'. Each node $s \in \mathcal{T}$ is associated with the sequence corresponding to the path from this node to the root. This sequence (also called a 'context') is also denoted by $s$. Each node maintains two counters $N_0$ and $N_1$ that count the number of '0's (resp. '1's) appearing after the contexts $s$ in the training sequence. We start with a perfect binary tree of height $D$ and set the counters in each node to zero. We process the training sequence $q_1^n$ by considering all $n - D$ contexts of size $D$ and for each context we update the counters of the nodes along the path defined by this context. Upon completion we can estimate the probability of a test sequence $x_1^T$ using the relation (8) by computing $\hat{P}_{\text{CTW}}^\epsilon(x_1^T)$. This probability estimate for the sequence $x_1^T$ induces the conditional distributions $\hat{P}_{\text{CTW}}(x_i|x_{i-D+1}^{i-1})$, as noted in Section 2. For example, in Figure 3 we depict the CTW tree with $D = 2$ constructed according to the training sequence $q_1^9 = 101011010$.

---

13. As shown by Tjalkens et al. (1997), the above simple form of the KT estimator is equivalent to the original form, given in terms of a Dirichlet distribution by Krichevsky and Trofimov (1981). The (original) KT-estimated probability of a sequence containing $N_0$ zeros and $N_1$ ones is $\int_0^1 \frac{(1-\theta)^{N_0}\theta^{N_1}}{\pi\sqrt{\theta(1-\theta)}}d\theta$.

14. There is a natural coding for $D$-bounded models such that the length of the code in bits for a model $M$ is exactly $C_D(M)$.

This straightforward implementation requires $O(|\Sigma|^D)$ space (and time) for learning and $O(T \cdot |\Sigma|^D)$ for predicting $x_1^T$. Clearly, such time and space complexities are hopeless even for moderate $D$ values. We presented the algorithm this way for simplicity. However, as already mentioned by Willems et al. (1995), it is possible to obtain linear time complexities for both training and prediction. The idea is to only construct tree paths that occur in the training sequence. The counter values corresponding to all nodes along unexplored paths remain zero and, therefore, the estimated CTW probabilities for entire unvisited subtrees can be easily computed (using closed-form formulas). The resulting time (and space) complexity for training is $O(Dn)$ and for prediction, $O(TD)$. More details on this efficient implementation are nicely explained by Sadakane et al. (2000) (see also Tjalkens & Willems, 1997).

| $s$ | $N_0$ | $N_1$ | $\hat{P}_{\text{KT}}^s(q)$ | $\hat{P}_{\text{CTW}}^s(q)$ |
|-----|-------|-------|------|------|
| $\epsilon$ | 3 | 4 | 63/7680 | 21/1024 |
| 0 | 0 | 3 | 21/64 | 21/64 |
| 1 | 3 | 1 | 7/160 | 3/80 |
| 00 | 0 | 0 | 1 | 1 |
| 10 | 0 | 3 | 21/64 | 21/64 |
| 01 | 2 | 1 | 1/16 | 1/16 |
| 11 | 1 | 0 | 1/2 | 1/2 |

Figure 3: A CTW tree ($D = 2$) corresponding to the training sequence $q_1^9 = 101011010$, where the first two symbols in $q$ ($q_1^2$) are neglected. We present, for each tree node, the values of the counters $N_0$, $N_1$ and the estimations $\hat{P}_{\text{KT}}(0|\cdot)$ and $\hat{P}_{\text{CTW}}(0|\cdot)$. For example, for the leaf 10: $N_0 = 0$ since 0 does not appear after 10 in $q_1^9 = 101011010$; 1 appears (in $q_1^9$) three times after 10, therefore, $N_1 = 3$; $\hat{P}_{\text{KT}}(0|10) = \frac{0+\frac{1}{2}}{0+3+1} = \frac{1}{8}$; because 10 is a leaf, according to Equation (8) $\hat{P}_{\text{KT}}(0|10) = \hat{P}_{\text{CTW}}(0|10)$.

## 3.4 CTW for Multi-Alphabets

The above CTW algorithm for a binary alphabet can be extended in a natural manner to handle sequences over larger alphabets. One must (i) extend the KT-estimator for larger alphabets; and (ii) extend the recurrence relation (8). While these extensions can be easily obtained, it is reported that the resulting algorithm preforms poorly (see Volf, 2002, chap. 4). As noted by Volf, the reason is that the extended KT-estimator does not provide efficient smoothing for large alphabets. Therefore, the problem of extending the CTW algorithm for large alphabets is challenging.

Several CTW extensions for larger alphabets have been proposed. Here we consider two. The first is a naive application of the standard binary CTW algorithm over a binary representation of the sequence. The binary representation is naturally obtained when the size of the alphabet is a power of 2. Suppose that $k = \log_2 |\sigma|$ is an integer. In this case we generate a binary sequence by concatenating binary words of size $k$, one for each alphabet symbol. If $\log_2 |\sigma|$ is not an integer we take $k = \lceil \log_2 |\sigma| \rceil$. We denote the resulting algorithm by BI-CTW. A more sophisticated binary decomposition of CTW was considered by Tjalkens et al. (1997). There eight binary machines were simultaneously constructed, one for each of the eight binary digits of the ascii representation of text. This decomposition does not achieve the best possible compression rates over the Calgary Corpus.

The second method we consider is Volf's 'decomposed CTW', denoted here by DE-CTW (Volf, 2002). The DE-CTW uses a tree-based hierarchical decomposition of the multi-valued prediction problem into binary problems. Each of the binary problems is solved via a slight variation of the binary CTW algorithm.

Let $\Sigma$ be an alphabet with size $k = |\Sigma|$. Consider a full binary tree $\mathcal{T}_\Sigma$ with $k$ leaves. Each leaf is uniquely associated with a symbol in $\Sigma$. Each internal node $v$ of $\mathcal{T}_\Sigma$ defines the binary problem of predicting whether the next symbol is a leaf on $v$'s left subtree or a leaf on $v$'s right subtree. For example, for $\Sigma = \{a,b,c,d,r\}$, Figure 4 depicts a tree $\mathcal{T}_\Sigma$ such that the root corresponds to the problem of predicting whether the next symbol is a or one of b,c,d and r. The idea is to learn a binary predictor, based on the CTW algorithm, for each internal node.

In a simplistic implementation of this idea, we construct a binary CTW for each internal node $v \in \mathcal{T}_\Sigma$. We project the training sequence over the "relevant" symbols (i.e., corresponding to the subtree rooted by $v$) and translate the symbols on $v$'s left (resp., right) sub-tree to 0s (resp., 1s). After training we predict the next symbol $\sigma$ by assigning each symbol a probability that is the product of binary predictions along the path from the root of $\mathcal{T}_\Sigma$ to the leaf labeled by $\sigma$.

Unfortunately, this simple approach may result in poor performance and a slight modification is required.[15] For each internal node $v \in \mathcal{T}_\Sigma$, let CTW$_v$ be the associated binary predictor designed to handle the alphabet $\Sigma_v \subseteq \Sigma$. Instead of translating the projected sequence (projected over $\Sigma_v$) to a binary sequence, as in the simple approach, we construct CTW$_v$ based on a $|\Sigma_v|$-ary context tree. Thus, CTW$_v$ still generates predictions over a binary alphabet but expands a suffix tree using the (not necessarily binary) $\Sigma_v$ alphabet. To generate binary predictions CTW$_v$ utilizes, in each node of the context tree, two counters for computing binary KT-estimations (as in the standard CTW algorithm). For example, in Figure 4(b) we depict CTW$_3$ whose binary problem is defined by $\mathcal{T}_\Sigma$ of Figure 4(a). Another modification suggested by Volf, is to use a variant on the KT-estimator. Volf shows that the estimator $P_e(0|q) = \frac{N_0(q) + \frac{1}{2}}{N_0(q) + N_1(q) + 1/8}$ achieves better results in practice. We used this estimator for the DE-CTW implementation.[16]

A major open issue when applying the DE-CTW algorithm is the construction of an effective decomposition tree $\mathcal{T}_\Sigma$. Following (Volf, 2002, Chapter 5), we employ the heuristic now detailed. Given a training sequence $q_1^n$, the algorithm takes $\mathcal{T}_\Sigma$ to be Hoffman's code-tree of $q_1^n$ (see Cover & Thomas, 1991, chapter 5.6) based on the frequency counts of the symbols in $q_1^n$.

---

15. Consider the following example. Let $\Sigma = \{a,b,z\}$ and consider a decomposition $\mathcal{T}_\Sigma$ with two internal nodes where the binary problem of the root discriminates between $\{a,b\}$ and $\{z\}$ (and the other internal node corresponds to distinguishing between a and b). Using the simplistic approach, if we observe the training sequence azaz $\cdots$ azazb, we will assign very high probability to z given the context zb, which is not necessarily supported by the training data.

16. We also used this estimator in the protein classification experiment.

The time and space complexity of the DE-CTW algorithm, based on the efficient implementation of the binary CTW, are $O(|\Sigma|Dn)$ for training and and $O(|\Sigma|DT)$ for prediction.

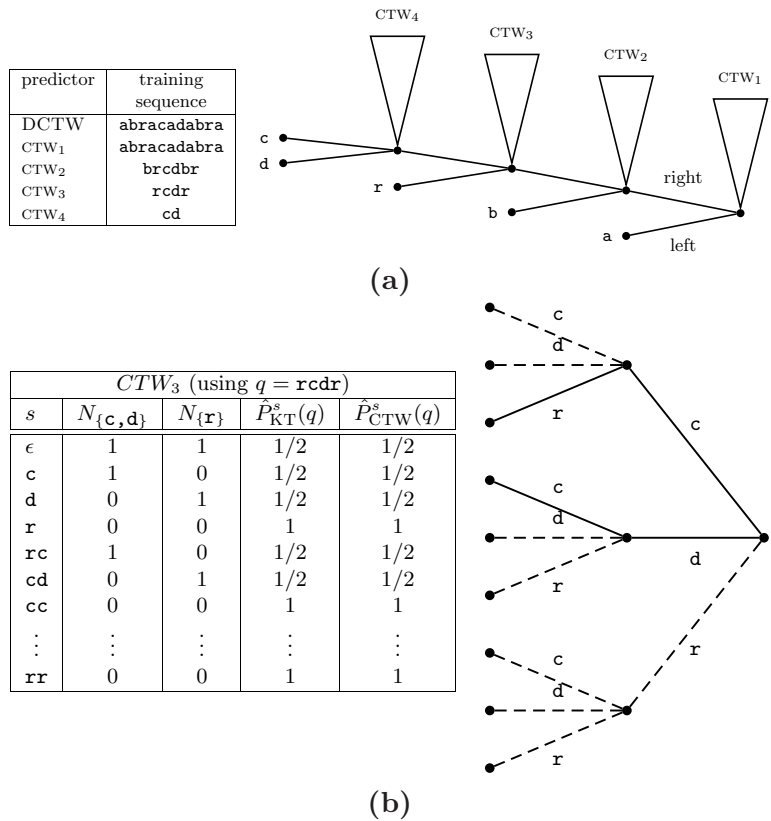

**(a)**

**(b)**

Figure 4: A DE-CTW tree corresponding to the training sequence $q_1^{11} = \texttt{abracadabra}$. (a) depicts the decomposition tree $\mathcal{T}$. Each internal node in $\mathcal{T}$ uses a CTW predictor to "solve" a binary problem. In (b) we depict CTW$_3$ whose binary problem is: "determine if $\sigma \in \{\texttt{c},\texttt{d}\}$ (or $\sigma = \texttt{r}$)". Tree paths of contexts that do not appear in the training sequence are marked with dashed lines.

## 3.5 Probabilistic Suffix Trees (PST)

The Probabilistic Suffix Tree (PST) prediction algorithm (Ron et al., 1996) attempts to construct the single "best" $D$-bounded VMM according to the training sequence. It is assumed that an upper bound $D$ on the Markov order of the "true source" is known to the learner.

A PST over $\Sigma$ is a non empty rooted tree, where the degree of each node varies between zero (for leaves) and $|\Sigma|$. Each edge in the tree is associated with a unique symbol in $\Sigma$. These edge labels define a unique sequence $s$ for each path from a node $v$ to the root. The sequence $s$ labels the node $v$. Any such PST tree induces a "suffix set" $S$ consisting of the

labels of all the nodes. The goal of the PST learning algorithm is to identify a good suffix set $S$ for a PST tree and to assign a probability distribution $\hat{P}(\sigma|s)$ over $\Sigma$, for each $s \in S$. Note that a PST tree may not be a tree source, as defined in Section 3.3. The reason is that the set $S$ is not necessarily proper. We describe the learning phase of the PST algorithm, which can be viewed as a three stage process.

1. First, the algorithm extracts from the training sequence $q_1^n$ a set of candidate contexts and forms a candidate "suffix set" $\hat{S}$. Each contiguous sub-sequence $s$ of $q_1^n$ of length $\leq D$, such that the frequency of $s$ in $q_1^n$ is larger than a user threshold will be in $\hat{S}$. Note that this construction guarantees that $\hat{S}$ can be accommodated in a (PST) tree (if a context $s$ appears in $\hat{S}$, then all suffixes of $s$ must also be in $\hat{S}$). For each candidate $s$ we associate the conditional (estimated) probability $\hat{P}(\sigma|s)$, based on a direct maximum likelihood estimate (i.e., a frequency count).

2. In the second stage, each $s$ in the candidate set $\hat{S}$ is examined using a two-condition test. If the context $s$ passes the test, $s$ and all its suffixes are included in the final PST tree. The test has two conditions that must hold simultaneously:

   (i) The context $s$ is "meaningful" for some symbol $\sigma$; that is, $\hat{P}(\sigma|s)$ is larger than a user threshold.

   (ii) The context $s$ contributes additional information in predicting $\sigma$ relative to its "parent"; if $s = \sigma_k\sigma_{k-1}\cdots\sigma_1$, then its "parent" is its longest suffix $s' = \sigma_{k-1}\cdots\sigma_1$. We require that the ratio $\frac{\hat{P}(\sigma|s)}{\hat{P}(\sigma|s')}$ be larger than a user threshold $r$ or smaller than $1/r$.

   Note that if a context $s$ passes the test there is no need to examine any of its suffixes (which are all added to the tree as well).

3. In the final stage, the probability distributions associated with the tree nodes (contexts) are smoothed. If $\hat{P}(\sigma|s) = 0$, then it is assigned a minimum probability (a user-defined constant) and the resulting conditional distribution is re-normalized.

Altogether the PST learning algorithm has five user parameters, which should be selected based on the learning problem at hand. The PST learning algorithm has a PAC style performance guarantee ensuring that with high probability the redundancy approaches zero at rate $1/n^{1/c}$ for some positive integer $c$.[17] An example of the PST tree constructed for the `abracadabra` training sequence is given in Figure 5.

The PST learning phase has a time complexity of $O(Dn^2)$ and a space complexity of $O(Dn)$. The time complexity for calculating $\hat{P}(\sigma|s)$ is $O(D)$.

### 3.6 LZ-MS: An Improved Lempel-Ziv Algorithm

There are plenty of variations on the classic LZ78 *compression* algorithm (see Bell et al., 1990, for numerous examples). Here we consider a recent variant of the LZ78 *prediction* algorithm due to Nisenson et al. (2003). The algorithm has two parameters $M$ and $S$ and, therefore, its acronym here is LZ-MS.

---

17. Specifically, the result is that, with high probability ($\geq 1 - \delta$), the Kullback-Leibler divergence between the PST estimated probability distribution and the true underlying (unknown) VMM distribution is not larger than $\varepsilon$ after observing a training sequence whose length is polynomial in $\frac{1}{\varepsilon}$, $\frac{1}{\delta}$, $D$, $|\Sigma|$ and the number of states in the unknown VMM model.

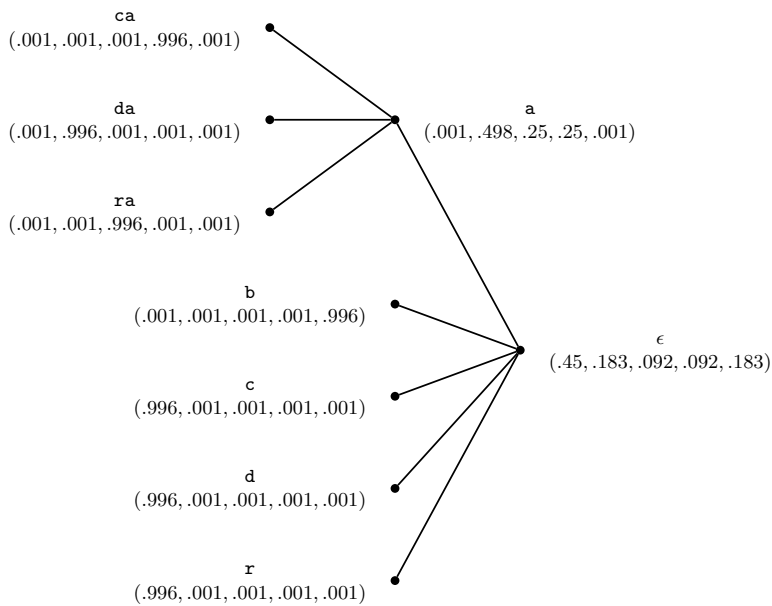

Figure 5: A PST tree corresponding to the training sequence $x_1^{11} = \texttt{abracadabra}$ and the parameters ($P_{min} = 0.001$, $\gamma_{min} = 0.001$, $\alpha = 0.01$, $r = 1.05$, $D = 12$). Each node is labelled with a sequence $s$ and the distribution ($\hat{P}_s(\texttt{a}), \hat{P}_s(\texttt{b}), \hat{P}_s(\texttt{c}), \hat{P}_s(\texttt{d}), \hat{P}_s(\texttt{r})$).

A major advantage of the LZ78 algorithm is its speed. This speed is made possible by compromising on the systematic counts of all sub-sequences. For example, in Figure 1, the sub-sequence $\texttt{br}$ is not parsed and, therefore, is not part of the tree. However, this sub-sequence is significant for calculating $\hat{P}(\texttt{a}|\texttt{br})$. While for a very long training sequence this compromise will not affect the prediction quality significantly, for short training sequences the LZ78 algorithm yields sparse and noisy statistics. Another disadvantage of the LZ78 algorithm is its loss of context when calculating $\hat{P}(\sigma|s)$. For example, in Figure 1, when taking $\sigma = \texttt{b}$ and $s = \texttt{raa}$, the LZ78 prediction algorithm will compute $\hat{P}(\texttt{b}|\texttt{raa})$ by traversing the tree along the $\texttt{raa}$ path (starting from the root) and will end on a leaf. Therefore, $\hat{P}(\texttt{b}|\texttt{raa}) = \hat{P}(\texttt{b}|\epsilon) = 5/33$. Nevertheless, we could use the suffix $\texttt{a}$ of $\texttt{raa}$ to get the more accurate value $\hat{P}(\texttt{b}|\texttt{a}) = 5/17$. These are two major deficiencies of LZ78. The LZ-MS algorithm attempts to overcome both these disadvantages by introducing two corresponding heuristics. This algorithm improves the LZ78 predictions by extracting more phrases during learning and by ensuring a minimal context for the next phrase, whenever possible.

The first modification is termed *input shifting*. It is controlled by the $S$ parameter and used to extract more phrases from the training sequence. The training sequence $q_1^n = q_1 q_2 \cdots q_n$ is parsed $S + 1$ times, where in the $i$th parsing the algorithm learns the sequence $q_i q_{i+1} \cdots q_n$ using the standard LZ78 learning procedure. The newly extracted phrases and counts are combined with the previous ones (using the same tree). Note that by taking $S = 0$ we leave the LZ78 learning algorithm intact. For example, the second row of Table 1 illustrates the effect of input shifting with $S = 1$, which clearly enriches the set of parsed phrases.

The second modification is termed *back-shift parsing* and is controlled by the $M$ parameter. Back-shift parsing attempts to guarantee a minimal context of $M$ symbols when calculating probabilities. In the learning phase the algorithm back-shifts $M$ symbols after each phrase extraction. To prevent the case where more symbols are back-shifted than parsed (which can happen when $M > 1$), the algorithm requires that back shifting remain within the last parsed phrase. For example, on the third row of Table 1 we see the resulting phrase set extracted from the training sequence using $M = 1$. Observe that after extracting the first phrase (a) LZ-MS back-shifts one symbol, therefore, the next parsed symbol is (again) 'a'. This implies that the next extracted phrase is ab. Since LZ-MS back-shifts after each phrase extraction we conclude with the resulting extracted phrase set.

The back-shift parsing mechanism may improve the calculation of conditional probabilities, which is also modified as follows. Instead of returning to the root after traversing to a leaf, the last $M$-traversed symbols are first traced down from the root to some node $v$, and then the new traversal begins from $v$ (if $v$ exists). Take, for example, the calculation of $\hat{P}(b|raa)$ using the tree in Figure 1. If we take $M = 0$, then we end with $\hat{P}(b|raa) = \hat{P}(b|\epsilon) = 5/33$; on the other hand, if we take $M = 1$, we end with $\hat{P}(b|raa) = \hat{P}(b|a) = 5/17$.

| LZ78$(M, S)$ | Phrases parsed from abracadabra |
|---|---|
| LZ78$(0, 0) =$ LZ78 | {a,b,r,ac,ad,ab,ra} |
| LZ78$(0, 1)$ | {a,b,r,ac,ad,ab,ra,br,aca,d,abr} |
| LZ78$(1, 0)$ | {a,ab,b,br,r,ra,ac,c,ca,ad,d,da,abr} |
| LZ78$(1, 1)$ | {a,ab,b,br,r,ra,ac,c,ca,ad,d,da,abr,bra,aca,ada,abra} |
| LZ78$(2, 0)$ | {a,ab,abr,b,br,bra,r,ra,rac,ac,aca,c,ca,cad,ad,ada,d,da, dab,abra} |
| LZ78$(2, 1)$ | {a,ab,abr,b,br,bra,r,ra,rac,ac,aca,c,ca,cad,ad,ada,d,da, dab,abra,brac,acad,adab} |
| LZ78$(2, 2)$ | {a,ab,abr,b,br,bra,r,ra,rac,ac,aca,c,ca,cad,ad,ada,d,da, dab,abra,brac,acad,adab,raca,cada,dabr} |

Table 1: Phrases parsed from $q_1^{11} =$ abracadabra by LZ-MS for different values of $M$ and $S$. Note that the phrases appear in the order of their parsing.

Both 'input shifting' and 'back-shift parsing' tend to enhance the statistics we extract from the training sequence. In addition, back-shift parsing tends to enhance the utilization of the extracted statistics. Table 1 shows the phrases parsed from $q_1^{11} = abracadabra$ by LZ-MS for several values of $M$ and $S$. The phrases appear in the order of their parsing. Observe that each of these heuristics can introduce a slight bias into the extracted statistics.

It is interesting to compare the relative advantage of input shifting and back-shift parsing. In Appendix C we provide such a comparison using a prediction simulation over the Calgary Corpus. The results indicate that back-shift on its own can provide more power to LZ78 than input shifting parsing alone. Nevertheless, the utilization of both mechanisms is always better than using either alone.

The time complexity of the LZ-MS learning algorithm is at most $MS$ times the complexity of LZ78. Prediction using LZ-MS can take at most $M$ times the prediction using LZ78.

## 4. Experimental Setup

We implemented and tested the six VMM algorithms described in Section 3 using discrete sequences from three domains: text, music and proteins (see more details below). The source code of Java implementations for the six algorithms is available at `http://www.cs.technion.ac.il/~ronbeg/vmm`. We considered the following prediction setting. The learner is given the first half of a sequence (e.g., a song) and is required to predict the rest. We denote the first half by $q_1^n = q_1 \cdots q_n$ and the second, by $x_1^n = x_1 \cdots x_n$. $q_1^n$ is called the *training sequence* and $x_1^n$ is called the *test sequence*. During this learning stage the learner constructs a predictor, which is given in terms of a conditional distribution $\hat{P}(\sigma|s)$. The predictor is then fixed and tested over $x_1^n$. The quality of the predictor is measured via the average log-loss of $\hat{P}(\cdot|\cdot)$ with respect to $x_1^n$ as given in Equation (1).[18] For the protein dataset we also considered a different, more natural setting in which the learner is provided with a number of training sequences emanating from some unknown source, representing some domain. The learner is then required to provide predictions for sequences from the same domain.

During training, the learner should utilize the training sequence for learning a probabilistic model, as well as for selecting the best possible hyper-parameters. Each algorithm selects its hyper-parameters from a cross product combination of "feasible" values, using a cross-validation (CV) scheme over the training sequence. In our experiments we used a five-fold CV. The training sequence $q_1^n$ was segmented into five (roughly) equal sized contiguous non-overlapping segments and each fold used one such segment for testing and a concatenation of rest of the segments for training. Before starting the experiments we selected, for each algorithm, a set of "feasible" values for its vector of hyper-parameters. Each possible vector (from this set) was tested using five-fold CV scheme (applied over the training sequence!) yielding five loss estimates for this vector. The parameter vector with the best median[19] (over the five estimates) was selected for training the algorithm over the entire training sequence. See Appendix A for more details on the exact choices of hyper-parameters for each of the algorithms. Note that our setting is different from the standard setup used when testing online lossless compression schemes, where the compressor starts processing the test sequence without any prior knowledge. Also, it is usually the case that compression algorithms do not optimize their hyper-parameters (e.g., for text compression PPM-C is usually applied with a fixed $D = 5$).

The sequences we consider belong to three different domains: English text, music pieces and proteins.

---

18. Cover and King (1978) considered a very similar prediction game in their well-known work on the estimation of the entropy of English text.

19. We used the median rather than average since the median of a small set of numbers is considerably more robust against outliers; see, e.g., `http://standards.nctm.org/document/eexamples/chap6/6.6/`

| Domain | # Sequences | $|\Sigma|$ | Sequence Lengths | | |
|--------|-------------|-----------|------|------|------|
| | | | Min | Avg | Max |
| Text | 18 | 256 | 11954 | 181560 | 785396 |
| Music | 285 | 256 | 24 | 12726 | 193662 |
| Protein | 19676 | 20 | 21 | 187 | 8599 |

Table 2: Some essential properties of the datasets. For the protein prediction setup we considered only sequences of size larger than 100.

- For the English text we chose the well-known 'Calgary Corpus' (Bell et al., 1990), which is traditionally used for benchmarking lossless compression algorithms.[20]

- The music set was assembled from MIDI files of music pieces.[21] The musical benchmark was compiled using a variety of well-known pieces of different styles. The styles we included are: classical, jazz and rock/pop. All the pieces we consider are polyphonic (played with several instruments simultaneously). Each MIDI file for a polyphonic piece consists of several channels (usually one channel per instrument). We considered each channel as an individual sequence; see Appendix B for more details.

- The protein set includes proteins (amino-acid sequences) from the well-known *Structural Classification of Proteins (SCOP)* database (Murzin et al., 1995). Here we used all sequences in release 1.63 of SCOP, after eliminating redundancy (i.e., only one copy was retained of sequences that are 100% identical). This set is hierarchically classified into classes according to biological considerations. We relied on this classification for testing classifiers constructed from the VMM algorithms over a number of large classification tasks (see Section 6).

Table 2 summarizes some statistics of the three benchmark tests. The three datasets can be obtained at `http://www.cs.technion.ac.il/~ronbeg/vmm`.

## 5. Prediction Results

Here we present the prediction results of the six VMM algorithms for the three domains. The performance is measured via the log-loss as discussed above. Recall that the average log-loss of a predictor (over some sequence) is a slightly optimistic estimate of the compression rate (bits/symbol) that could be achieved by the algorithm over that sequence.

Table 3 summarizes the results for the textual domain (Calgary Corpus). We see that DE-CTW is the overall winner with an average loss of 3.02. The runner-up is PPM-C with an average loss of 3.03. However, there is no statistically significant difference between the two, as indicated by standard error of the mean (SEM) values. The worst algorithm is the LZ78 with an average loss of 3.76. We see that in most cases DE-CTW and PPM-C share the first and second best results, LZ-MS is a runner-up in five cases, and LZ78 wins one case and is a runner-up in one case. PST is the winner in a single case.

---

20. The Calgary Corpus is available at `ftp://ftp.cpsc.ucalgary.ca/pub/projects/text.compression.corpus`. Note that this corpus contains, in addition to text, also a few binary files and some source code of computer programs, written in a number of programming languages.

21. MIDI (= *Musical Instrument Digital Interface*) is a standard protocol and language for electronic musical instruments. A standard MIDI file includes instructions on which notes to play and how long and loud to play them.

It is interesting to compare the loss results of Table 3 to known compression results of (some of) the algorithms for the same corpus, which are 2.14 bps for DE-CTW and 2.48 bps for PPM-C (see Volf, 2002; Cleary & Teahan, 1997, respectively). While these results are dramatically better than the average log-loss we obtain for these algorithms, note that the setting is considerably different and in the compression applications the algorithms are continually learning the sequence. A closer inspection of this disparity between the compression results and our training/test results reveals that the difference is in fact not that large. Specifically, the large average log-loss is mainly due to one of the Calgary Corpus files, namely the *obj1* file for which the log-loss of PPM-C is 6.75, while the compression results of Cleary and Teahan (1997) are 3.76 bps. Without *obj1* file the average log-loss obtained by PPM-C on the Calgary Corpus is 2.81. Moreover, the 2.48 result is obtained on a smaller corpus that does not include the four files *paper3-6*. Without these files the average log-loss of PPM-C is 2.65. This small difference may indicate that the involved sequences are not governed by a stationary distribution. Notice also that the compression results indicate that DE-CTW is significantly better than PPM-C. However, in our setting these algorithms perform very similarly.

Next we present the results for the music MIDI files, which are summarized in Table 4. The table has four sections corresponding to classical pieces, jazz, 14 improvisations of the same jazz piece and rock/pop. Individual average losses for these sections are specified as well. DE-CTW is the overall winner and PPM-C is the overall runner-up. Both these algorithms are significantly better than the rest. The worst algorithm is, again, LZ78.

Some observations can be drawn about the "complexity" of the various pieces by considering the results of the best predictors with respect to the individual sequences. Taking the DE-CTW average losses, in the classical music domain we see that the Mozart pieces are the easiest to predict (with an average loss of at most 0.93). The least predictable pieces are Debussy's Children Corner 6 (1.86) and Beethoven's Symphony 6(1.78 loss). The most predictable piece overall is the rock piece "You really got me" by the Kinks. The least predictable rock piece is "Sunshine of your love" by Cream. The least predictable piece overall is the jazz piece "Satin doll 2" by Duke Ellington. The most predictable jazz piece is "The girl from Ipanema" by Jobim. Among the improvisations of "All of me", there is considerable variability starting with 0.5 loss (All of me 12) and ending with 1.65 loss (All of me 8). We emphasize that such observations may have a qualitative value due to the great variability among different arrangements of the same musical piece (particularly jazz tunes involving improvisation). The readers are encouraged to listen to these pieces and judge the "complexity" of these pieces for themselves. MIDI files of all the pieces are available at `http://www.cs.technion.ac.il/~ronbeg/vmm`.

The prediction results for the protein corpus appear in Table 5. Here PPM-C is the overall winner and DE-CTW is the runner-up. The difference between them and the other algorithms is not very large, with the exception of the PST algorithm, which suffered a significantly larger average loss.[22]

| Sequence (length·$10^3$) | BI-CTW | DE-CTW | LZ78 | LZ-MS | PPM-C | PST |
|---|---|---|---|---|---|---|
| *bib*(111) | 2.15 | **1.8\*** | 3.25 | 2.26 | **1.91** | 2.57 |
| *book*1(785) | 2.29 | **2.2\*** | 3.3 | 2.57 | **2.27** | 2.5 |
| *book*2(611) | 2.5 | **2.36\*** | 3.4 | 2.63 | **2.38** | 2.77 |
| *geo*(102) | 4.93 | **4.47\*** | **4.48** | **4.48** | 4.62 | 4.86 |
| *news*(377) | 3.04 | **2.81\*** | 3.72 | 3.07 | **2.82** | 3.4 |
| *obj*1(22) | 7 | 6.82 | **5.94\*** | **6** | 6.75 | 6.37 |
| *obj*2(247) | 3.92 | 3.56 | 3.99 | **3.48** | **3.45\*** | 3.72 |
| *paper*1(53) | 3.48 | **2.95\*** | 4.08 | 3.38 | **3.08** | 3.93 |
| *paper*2(82) | 2.75 | **2.49\*** | 3.63 | 2.81 | **2.53** | 3.19 |
| *paper*3(47) | 2.96 | **2.59\*** | 3.82 | 3.09 | **2.75** | 3.4 |
| *paper*4(13) | 3.87 | 3.63 | 4.13 | **3.61** | **3.36\*** | 3.99 |
| *paper*5(12) | 4.57 | 4.5 | 4.51 | **4.16** | **3.92\*** | 4.6 |
| *paper*6(38) | 3.9 | **3.21\*** | 4.22 | 3.6 | **3.31** | 4 |
| *pic*(513) | 0.75 | **0.71** | 0.78 | 0.79 | 0.73 | **0.7\*** |
| *progc*(40) | 3.28 | **2.85\*** | 3.95 | 3.22 | **2.94** | 3.71 |
| *progl*(72) | 3.2 | **2.85\*** | 3.7 | 3.26 | **2.99** | 3.66 |
| *progp*(49) | 2.88 | **2.53** | 3.24 | 2.65 | **2.52\*** | 2.95 |
| *trans*(94) | 2.65 | **2.08\*** | 3.56 | 2.61 | **2.29** | 3.01 |
| Average±SEM | $3.34 \pm 0.07$ | $\mathbf{3.02 \pm 0.07*}$ | $3.76 \pm 0.05$ | $3.2 \pm 0.06$ | $\mathbf{3.03 \pm 0.07}$ | $3.52 \pm 0.06$ |

Table 3: Average log-loss (equivalent to bits/symbol) of the VMM algorithms over the Calgary Corpus. For each sequence (row) the winner is starred and appears in boldface. The runner-up appears in boldface.

The striking observation, however, is that none of the algorithms could beat a trivial prediction based on a (zero-order) "background" distribution of amino acid frequencies as measured over a large database of proteins. The entropy of this background distribution is 4.19. Moreover, the entropy of a yet more trivial predictor based on the uniform distribution over an alphabet of size 20 is $4.32 \approx \log_2(20)$. Thus, not only could the algorithms not outperform these two trivial predictors, they generated predictions that are sometimes considerably worse.

These results indicate that the first half of a protein sequence does not contain predictive information about its second half. This is consistent with the general perception that protein chains are not repetitive [23]. Rather, usually they are composed of several different elements called domains, each one with own specific function and unique source distribution (Rose,

---

22. The failure of PST in this setting could be a result of an inadequate set of possible values for its hyper-parameters; see Appendix A for details on the choices made.

23. This is not always true. It is not unusual to observe similar patterns (though not identical) within the same protein. However, this phenomenon is observed for a relatively small number of proteins.

| Sequence | BI-CTW | DE-CTW | LZ78 | LZ-MS | PPM-C | PST |
|---|---|---|---|---|---|---|
| Goldberg Variations | 1.09 | **1*** | 1.59 | 1.28 | **1.04** | 1.15 |
| Toccata and Fuga | 1.17 | **1.08*** | 1.67 | 1.34 | **1.14** | 1.37 |
| for Elise | 1.76 | **1.57** | 2.32 | 1.83 | **1.57** | 2.08 |
| Beethoven - Symphony 6 | 1.91 | **1.78*** | 2.2 | 1.99 | **1.79** | 2.26 |
| Chopin - Etude 1 op. 10 | 1.14 | **1.07*** | 1.75 | 1.36 | **1.09** | 1.24 |
| Chopin - Etude 12 op. 10 | **1.67** | **1.54*** | 2.11 | 1.93 | 1.73 | 1.72 |
| Children's Corner - 1 | **1.28** | **1.23*** | 1.53 | 1.52 | 1.3 | 1.39 |
| Children's Corner - 6 | 2.03 | **1.86*** | 2.28 | 2.13 | **1.97** | 2.28 |
| Mozart K.551 | 1.04 | **0.85*** | 1.64 | 1.18 | **0.96** | 1.45 |
| Mozart K.183 | 1.04 | **0.93*** | 1.71 | 1.2 | **0.99** | 1.7 |
| Rachmaninov Piano Concerto 2 | 1.16 | **1.06*** | 1.8 | 1.35 | **1.13** | 1.34 |
| Rachmaninov Piano Concerto 3 | 1.91 | 1.75 | 2.09 | 1.98 | 1.82 | 2.1 |
| **Average Classical** | $1.43 \pm 0.02$ | $1.31 \pm 0.02$ | $1.89 \pm 0.02$ | $1.59 \pm 0.02$ | $1.38 \pm 0.02$ | $1.67 \pm 0.03$ |
| Giant Steps | 1.51 | 1.33 | 2.01 | **1.29** | **1.24*** | 1.82 |
| Satin Doll 1 | 1.7 | **1.41*** | 2.28 | 1.73 | **1.47** | 1.89 |
| Satin Doll 2 | 2.56 | **2.27*** | 2.53 | 2.54 | **2.38** | 3.32 |
| The Girl from Ipanema | 1.4 | **1.11*** | 1.95 | 1.5 | **1.2** | 1.36 |
| 7 Steps to Heaven | 1.7 | **1.42*** | 2.1 | 1.87 | **1.53** | 1.86 |
| Stolen Moments | 1.81 | **1.24*** | 2.28 | 1.59 | **1.37** | 1.8 |
| **Average Jazz** | $1.78 \pm 0.06$ | $1.46 \pm 0.06*$ | $2.19 \pm 0.04$ | $1.76 \pm 0.04$ | $1.53 \pm 0.06$ | $2.01 \pm 0.09$ |
| All of Me 1 | 1.97 | **1.43*** | 2.33 | 1.97 | **1.73** | 1.88 |
| All of Me 2 | 2.51 | **1.05*** | 2.52 | 1.9 | 1.65 | **1.28** |
| All of Me 3 | 1.29 | **1.08*** | 1.97 | 1.42 | **1.17** | 1.45 |
| All of Me 4 | 1.83 | **1.4*** | 2.12 | 1.88 | **1.46** | 2.32 |
| All of Me 5 | 0.92 | **0.78*** | 1.72 | 1.15 | **0.9** | 0.96 |
| All of Me 6 | 1.62 | **1.21*** | 2.21 | 1.55 | **1.28** | 1.52 |
| All of Me 7 | 1.97 | **1.65*** | 2.35 | 1.94 | **1.75** | 2.44 |
| All of Me 8 | 1.88 | **1.61*** | 2.29 | 1.97 | **1.67** | 2.21 |
| All of Me 9 | 1.79 | **1.58*** | 2.2 | 1.97 | **1.7** | 2.13 |
| All of Me 10 | 1.32 | **1.07*** | 1.96 | 1.42 | **1.16** | 1.4 |
| All of Me 11 | **1.61** | **1.54*** | 2.26 | 1.93 | 1.72 | 1.68 |
| All of Me 12 | 0.79 | **0.5*** | 1.8 | 0.92 | **0.54** | 0.57 |
| All of Me 13 | 0.97 | **0.68*** | 1.78 | 1.12 | **0.7** | 0.83 |
| All of Me 14 | 1.85 | **1.63*** | 2.2 | 1.89 | **1.68** | 2.17 |
| **Average Jazz Impro.** | $1.59 \pm 0.17$ | $1.23 \pm 0.09*$ | $2.12 \pm 0.09$ | $1.65 \pm 0.11$ | $1.37 \pm 0.12$ | $1.63 \pm 0.15$ |
| Don't Stop till You Get Enough | 1.03 | **0.64*** | 1.82 | 1.07 | **0.69** | 0.72 |
| Sir Duke | 1.14 | **0.47** | 2 | 0.99 | 0.59 | **0.35*** |
| Let's Dance | 1.22 | **0.94*** | 1.95 | 1.33 | **1.06** | 1.28 |
| Like a Rolling Stone | 1.63 | **1.45*** | 2.07 | 1.7 | **1.48** | 1.83 |
| Sunshine of Your Love | 1.68 | **1.47*** | 2.16 | 1.81 | **1.58** | 2.01 |
| You Really Got Me | 0.31 | **0.16*** | 1 | 0.49 | 0.23 | **0.17** |
| **Average Rock/Pop** | $1.17 \pm 0.05$ | $0.85 \pm 0.04*$ | $1.83 \pm 0.04$ | $1.23 \pm 0.04$ | $0.94 \pm 0.04$ | $1.06 \pm 0.06$ |
| Average$\pm$SEM | $1.49 \pm 0.08$ | $1.21 \pm 0.05*$ | $2.00 \pm 0.05$ | $1.56 \pm 0.05$ | $1.30 \pm 0.06$ | $1.59 \pm 0.08$ |

Table 4: Music MIDI sequences. Average log-loss (equivalent to bits/symbol) of the VMM algorithms over the music corpus. The table is partitioned into four sections: classical pieces, jazz pieces, 14 improvisations of the same jazz piece "All of me" and rock pieces. Each number is an average loss corresponding to several MIDI channels (see Appendix B). For each sequence (row) the winner is starred and appears in boldface. The runner-up appears in boldface.

1979; Lesk & Rose, 1981; Holm & Sander, 1994). These unique distributions are usually more statistically different from each other (in terms of the KL-divergence) than from the background or uniform distribution. Therefore, a model that is trained on the first half of a protein is expected to perform worse than the trivial predictor - which explains the slight increase in the average code length. A similar finding was observed by Bejerano and Yona (2001) where models that were trained over specific protein families coded unrelated non-member proteins using an average code length that was higher than the entropy of the background distribution. This is also supported by other papers that suggest that protein sequences are incompressible (e.g. Nevill-Manning & Witten, 1999).

| Sequences Class | BI-CTW | DE-CTW | LZ78 | LZ-MS | PPM-C | PST |
|---|---|---|---|---|---|---|
| A (1310) | $4.85 \pm 0.003$ | $\mathbf{4.52 \pm 0.007}$ | $4.80 \pm 0.005$ | $4.79 \pm 0.007$ | $\mathbf{4.45 \pm 0.005*}$ | $8.07 \pm 0.01$ |
| B (1604) | $4.87 \pm 0.003$ | $\mathbf{4.55 \pm 0.006}$ | $4.83 \pm 0.005$ | $4.85 \pm 0.007$ | $\mathbf{4.47 \pm 0.005*}$ | $8.06 \pm 0.009$ |
| C (3991) | $4.85 \pm 0.002$ | $\mathbf{4.49 \pm 0.003}$ | $4.89 \pm 0.003$ | $4.92 \pm 0.004$ | $\mathbf{4.43 \pm 0.002*}$ | $8.05 \pm 0.005$ |
| D (2325) | $4.85 \pm 0.003$ | $\mathbf{4.65 \pm 0.006}$ | $4.91 \pm 0.004$ | $4.98 \pm 0.005$ | $\mathbf{4.54 \pm 0.004*}$ | $8.13 \pm 0.008$ |
| E (402) | $4.88 \pm 0.005$ | $\mathbf{4.36 \pm 0.006}$ | $4.78 \pm 0.009$ | $4.83 \pm 0.008$ | $\mathbf{4.33 \pm 0.004*}$ | $7.92 \pm 0.01$ |
| F (157) | $4.94 \pm 0.015$ | $\mathbf{4.37 \pm 0.017}$ | $4.71 \pm 0.014$ | $4.78 \pm 0.017$ | $\mathbf{4.34 \pm 0.013*}$ | $8.00 \pm 0.023$ |
| G (16) | $\mathbf{4.79 \pm 0.018*}$ | $5.00 \pm 0.052$ | $5.08 \pm 0.075$ | $5.03 \pm 0.105$ | $\mathbf{4.82 \pm 0.047}$ | $8.45 \pm 0.13$ |
| Average$\pm$SEM | $4.86 \pm 0.007$ | $\mathbf{4.56 \pm 0.014}$ | $4.85 \pm 0.16$ | $4.88 \pm 0.022$ | $\mathbf{4.48 \pm 0.011*}$ | $8.09 \pm 0.028$ |

Table 5: Protein sequences average prediction loss. A trivial distribution assignment based on a "background" amino-acid frequency (measured over a large protein database) has an entropy of 4.19. Note also that maximal entropy of any distribution is bounded above by the entropy of the uniform distribution $4.32 \approx \log_2(20)$. For each sequence (row) the winner is starred and appears in boldface. The runner-up appears in boldface. The last row represents the averages and standard error of the means.

## 6. Protein Classification

For protein sequences, a more natural setup than the sequence prediction setting is the classification setting. Biologists classify proteins into relational sets such as 'families', 'superfamilies', 'fold families' and 'classes'; these terms were coined by Dayhoff (1976) and Levitt and Chothia (1976). Being able to classify a new protein in its proper group can help to elucidate relationships between novel genes and existing proteins and characterize unknown genes.

Prediction algorithms can be used for classification using a winner-takes-all (WTA) approach. Consider a training set of sequences belonging to $k$ classes $C_1, \ldots, C_k$. For each class $C_i$ we train a sequence predictor $\hat{P}_i$. Given a test sequence $x$ we classify it as class $c = \arg\max_i \hat{P}_i(x)$.

We tested the WTA classifiers derived from the above VMM algorithms over several large protein classification problems. These problems consider the same protein sequences mentioned in the above prediction experiment. For classification we used the partition into subsets given in the SCOP database. Specifically, the SCOP database (Murzin et al., 1995) classifies proteins into a hierarchy of 'families', 'superfamilies', 'folds' and 'classes' where each 'class' is composed of several 'folds', each 'fold' is composed of several 'superfamilies' and each 'superfamily' is composed of several 'families'. This classification was done manually based on both sequence and structural similarities, resulting in seven major 'classes' (at the top level), 765 'folds',[24] 1231 'superfamilies' and 2163 'families' (release 1.63).

The seven major SCOP classes are quite distinct and can be easily distinguished (Kumarevel et al., 2000; Jin et al., 2003). Here we focused on the next level in this hierarchy, the 'fold' level. Our task was to classify proteins, that belong to the same 'class', into their unique 'fold'. Since there are seven 'classes' in the SCOP database, we have seven independent classification problems. The number of 'folds' (i.e., machine learning classes) in each 'class' appear in Table 6. For example, the protein 'class' D contains 111 'folds'. The classification problem corresponding to D is, therefore, a multi-class problem with 111 (machine learning) classes. Consider one (of the seven) classification problems. For each 'fold', a statistical model was trained from a training subset of sequences that belong to that 'fold'. Thus, the number of models we learn is the number of 'folds' (111 in the D 'Class'). The resulting classifier is a WTA combination of all these models. The test set for this classifier contains all the proteins from the same SCOP 'class' that do not appear in any training set.

Of the 765 'folds', we considered only 'folds' with at least 10 members. VMM models were trained (one for each 'fold') and tested. The test results are grouped and averaged per SCOP 'class'. The number of 'folds' in these classes varies between 12-111 (with an average of 58 'folds').

As before, the experiment was repeated five times using five-fold CV. We measured the performance using the average over the zero-one loss (zero loss for a correct classification). Within each fold we fine-tuned the hyper-parameters of the algorithm using the same "internal" five-fold cross validation methodology as described in Section 4.

In this setup we used a slightly different PST variant (denoted as PST∗) based on (Bejerano & Yona, 2001).[25] The PST∗ variant is basically a standard PST with the following two modifications: (i) A context is chosen to be a candidate according to the number of times it appears in different 'fold' sequences; (ii) A context $s$ is added to the tree if some symbol appears after $s$ at least some predefined number of times.

We present the classification-experiment results in Table 6. The consistent winner here is the LZ-MS algorithm, which came first in all of the classification problems, differing substantially from the runner-ups, DE-CTW and PPM-C.

In general, all the VMM algorithms' classifications had a good average success rate ($1-$ error) that ranges between 76% (for PST∗) to 86% (for LZ-MS). The latter result is quite surprising, considering that the class definition is based on structural properties that are not easily recognizable from the sequence. In fact, existing methods of sequence comparison usually fail to identify relationships between sequences that belong to different superfamilies within the same fold. The success rate of the LZ-MS algorithm in this task suggests that the algorithm might be very effective in other protein classification tasks.

---

24. Note the difference between the SCOP 'classes' and the term class we use in classification problems; similarly, note the difference between the SCOP 'folds' (that are subsets of the SCOP 'classes') from the folds we use in cross-validation. Throughout our discussion, the biological 'classes' and 'folds' are cited.

25. The results of the original PST algorithm were poor (probably due to an inadequate set of values for the hyper-parameters). Recall that the PST∗ (and all other algorithms) hyper-parameter values were chosen before starting the experiments, following the description given in Section 4.

| Protein 'Class' | # 'folds' | BI-CTW | DE-CTW | LZ78 | LZ-MS | PPM-C | PST* |
|---|---|---|---|---|---|---|---|
| $A$ | 86 | $0.26 \pm 0.035$ | $0.19 \pm 0.031$ | $0.28 \pm 0.035$ | $\mathbf{0.16 \pm 0.031}$ | $0.21 \pm 0.031$ | $0.22 \pm 0.009$ |
| B | 75 | $0.23 \pm 0.031$ | $0.17 \pm 0.031$ | $0.24 \pm 0.040$ | $\mathbf{0.14 \pm 0.031}$ | $0.18 \pm 0.035$ | - |
| C | 78 | $0.27 \pm 0.031$ | $0.22 \pm 0.031$ | $0.27 \pm 0.040$ | $\mathbf{0.17 \pm 0.035}$ | $0.21 \pm 0.031$ | - |
| D | 111 | $0.24 \pm 0.040$ | $0.19 \pm 0.031$ | $0.26 \pm 0.040$ | $\mathbf{0.16 \pm 0.031}$ | $0.18 \pm 0.035$ | $0.20 \pm 0.03$ |
| E | 17 | $0.12 \pm 0.031$ | $0.09 \pm 0.031$ | $0.1 \pm 0.044$ | $\mathbf{0.05 \pm 0.017}$ | $0.08 \pm 0.017$ | $0.16 \pm 0.076$ |
| F | 12 | $0.26 \pm 0.031$ | $0.21 \pm 0.044$ | $0.19 \pm 0.031$ | $\mathbf{0.17 \pm 0.03}$ | $0.24 \pm 0.067$ | $0.24 \pm 0.062$ |
| G | 29 | $0.33 \pm 0.044$ | $0.25 \pm 0.035$ | $0.31 \pm 0.05$ | $\mathbf{0.16 \pm 0.017}$ | $0.23 \pm 0.035$ | $0.36 \pm 0.044$ |
| Average | | $0.24$ | $0.18$ | $0.23$ | $\mathbf{0.14}$ | $0.19$ | $0.24$ |
| W. Average | | $0.25$ | $0.19$ | $0.25$ | $\mathbf{0.15}$ | $0.19$ | $0.28$ |

Table 6: Protein corpus results – classification error rate±SEM. five-fold cross validation results. Best results appear in boldface. W. Average stands for "weighted average" where the weights are the relative sizes of the protein classes.

The success of the algorithms in the classification setup contrasts with their poor protein prediction rates (given in Table 5). To investigate this phenomenon, we conducted another protein prediction experiment based on the classification setup. In each fold in this experiment (among the five-CV), a predictor $\hat{P}_i$ was trained using the concatenation of all training sequences in class $C_i$ and was tested against each individual sequence in the test of class $C_i$. The results of this prediction experiment are given in Table 7. Here we can see that all of the algorithms yield significantly smaller losses than the loss of the trivial protein "background" distribution. Here again, PPM and DE-CTW are favorites with significant statistical difference between them.

## 7. Related work

In this section we briefly discuss some results that are related to the present work. We restrict ourselves to comparative studies that include algorithms we consider here and to discussions of some recent extensions of these algorithms. We also discuss some results concerning compression and prediction of proteins.

The PPM idea and its overwhelming success in benchmark tests has attracted considerable attention and numerous PPM variants have been proposed and studied. PPM-D (Howard, 1993) is very similar to PPM-C, with a slight modification of the escape mechanism; see Section 3.2. The PPM* variant eliminates the need to specify a maximal order (Cleary & Teahan, 1997). For predicting the next symbol's probability, PPM* uses "deterministic" contexts, contexts that give probability one to some symbols. If no such context exists, then PPM* use the longest possible context (i.e., the original PPM strategy). One of the most successful PPM variants over the Calgary Corpus is the PPM-Z variant of Bloom (1998). The PPM-Z variant introduces two improvements to the PPM algorithm: (i) estimating the best model order, for each encoded symbol; (ii) adding a dynamic escape estimation for different contexts. To date, the most successful PPM variant is the 'complicated PPM with information-inheritance' (PPM-II) of Shkarin (2002), which achieves a compression rate of 2.041. Currently, this is the lowest published rate over the Calgary Corpus. The PPM-II algorithm relies on an improved escape mechanism similar to the one used by PPM-Z together with a different counting heuristic that aims to reduce the statistical sparseness that occurs in large Markovian contexts.

| Sequences Class | BI-CTW | DE-CTW | LZ78 | LZ-MS | PPM-C | PST* |
|---|---|---|---|---|---|---|
| A | $2.74 \pm 0.043$ | $\mathbf{1.95 \pm 0.033*}$ | $3.94 \pm 0.015$ | $2.29 \pm 0.034$ | $\mathbf{1.96 \pm 0.036}$ | $2.97 \pm 0.049$ |
| B | $2.57 \pm 0.033$ | $\mathbf{2.14 \pm 0.024}$ | $3.76 \pm 0.013$ | $2.18 \pm 0.026$ | $\mathbf{1.92 \pm 0.028*}$ | - |
| C | $2.63 \pm 0.036$ | $\mathbf{2.07 \pm 0.028}$ | $4.10 \pm 0.009$ | $2.30 \pm 0.031$ | $\mathbf{1.93 \pm 0.030*}$ | - |
| D | $2.59 \pm 0.038$ | $\mathbf{1.59 \pm 0.032*}$ | $3.65 \pm 0.021$ | $2.15 \pm 0.031$ | $\mathbf{1.83 \pm 0.031}$ | $2.64 \pm 0.037$ |
| E | $1.52 \pm 0.105$ | $\mathbf{1.24 \pm 0.081}$ | $3.47 \pm 0.052$ | $1.49 \pm 0.075$ | $\mathbf{1.16 \pm 0.081*}$ | $2.20 \pm 0.105$ |
| F | $2.99 \pm 0.155$ | $\mathbf{2.34 \pm 0.121}$ | $3.92 \pm 0.054$ | $2.49 \pm 0.13$ | $\mathbf{2.15 \pm 0.135*}$ | $3.04 \pm 0.141$ |
| G | $3.49 \pm 0.057$ | $\mathbf{2.30 \pm 0.057*}$ | $4.04 \pm 0.025$ | $2.88 \pm 0.058$ | $\mathbf{2.53 \pm 0.057}$ | $3.56 \pm 0.051$ |
| Average±SEM | $2.64 \pm 0.066$ | $\mathbf{1.94 \pm 0.053}$ | $3.84 \pm 0.027$ | $2.25 \pm 0.055$ | $\mathbf{1.92 \pm 0.056*}$ | $2.88 \pm 0.076$ |

Table 7: Protein prediction based on the classification setup. The classes B and C results of PST are missing. We terminated these runs after a week of computation. Observe the difference between these and the results of Table 5. Here all of the algorithms beat the two trivial predictors ("background" frequency and the uniform distribution) mentioned in caption of Table 5.

Traditionally, the comparison between PPM variants (and other lossless compression algorithms) is over the Calgary Corpus. Other corpora for lossless compression were proposed and are available. Two examples are the Canterbury Corpus and the 'Large Canterbury Corpus' (Arnold & Bell, 1997) and the Silesia Compression Corpus (Deorowicz, 2003), which contains significantly larger files than both the Calgary and Canterbury corpora.[26]

Bunton (1997) provides a (Calgary Corpus) comparison between PPM-C, PPM-D and PPM* (with its 'C' and 'D' versions). The average compression rates achieved by these algorithm are: 2.417, 2.448, 2.339 and 2.374, respectively, with a fixed order bound $D = 9$ for PPM-C and PPM-D. According to Bloom[27], his latest version of PPM-Z achieves an average rate of 2.086. Bunton also proposes various improvements for the PPM schemes. Her best improved PPM achieves a 2.177 rate.[28]

The general-purpose PPM scheme turns out to be rather flexible and allows adaptations to particular domains. For example, Nevill-Manning and Witten (1999) achieve a slightly better rate than the trivial $(\log_2(20))$ one for proteins, using a specialized variant of PPM-D that uses prior amino-acids frequencies to enhance the estimation. By combining domain knowledge into PPM using preprocessing and PPM that predicts over dual-alphabets, Drinic and Kirovski (2002) develop a specialized algorithm for compressing computer executable files (.exe). A PPM variant for compressing XML files was proposed by Cheney (2001). A text-mining utilization of the PPM algorithm was tested by Witten et al. (1999) and a few text classification setups were studied by Thaper (2001), Teahan and Harper (2001).

---

26. These corpora are available at `http://corpus.canterbury.ac.nz/` and `http://sun.iinf.polsl.gliwice.pl/~sdeor/corpus.htm`, respectively.
27. See `http://www.cbloom.com/src/ppmz.html`
28. At the time (1996) this variant was probably the most successful PPM version.

Context Tree Weighting (CTW), with its proven redundancy bound, has also attracted much attention. However, unlike the PPM approach, there are not as many empirical studies of this algorithm. A possible reason is that the original binary CTW (and straightforward extensions to larger alphabets) did not achieve the best lossless compression results (Tjalkens et al., 1997). The work of Volf, which extends the binary CTW to general alphabets, resulting in the DE-CTW algorithm (see Section 3.4), showed that CTW can achieve excellent compression rates. Since then, the CTW scheme has been extended in several directions. Perhaps the most important extension is a modified CTW scheme that does not require the maximal order hyper-parameter ($D$). The redundancy of this modified algorithm is bounded in a similar manner as the original version. This more recent CTW variant is thus universal with respect to the class of stationary and ergodic Markov sources of finite order.

Specialized CTW variants were proposed for various domains. For text, Aberg and Shtarkov (1997) proposed to replace the PPM estimator with the KT-estimator in the CTW model. They show that this combination of PPM-D and CTW outperforms PPM-D. This idea was examined by Sadakane et al. (2000) for both text and biological sequences. Their results also support the observation that the combination of a PPM predictor with the CTW model outperforms PPM-D compression both for textual data as well as for compressing DNA.

We are familiar with a number of extensions of the PST scheme. Singer (1997) extends the prediction scheme for sequence transduction and also utilizes the CTW model averaging idea. Bejerano and Yona (2001) incorporated biological prior knowledge into PST prediction and showed that the resulting PST-based similarity measure can outperform standard methods such as Gapped-BLAST, and is almost as sensitive as a hidden Markov model that is trained from a multiple alignment of the input sequences, while being much faster. Finally, Kermorvant and Dupont (2002) propose to improve the smoothing mechanism used in the original PST algorithm. Experiments with a protein problem indicate that the proposed smoothing method can help.

As mentioned earlier, there are many extensions of the LZ78 algorithm. See (Nelson & Gailly, 1996) for more details. It is interesting to note that Volf considered an "ensemble" consisting of both DE-CTW and PPM. He shows that the resulting algorithm performs nearly as well as the PPM-Z on the Calgary Corpus while beating the PPM-Z on the Canterbury Corpus.

Previous studies on protein classification used a variety of methods, including generative models such as HMMs (Sonnhammer et al., 1997), PSTs (Bejerano & Yona, 2001), and discriminative models such as Support Vector Machines (Ding & Dubchak, 2001; Leslie et al., 2004). Jaakkola et al. (1999) describe a model that combines the SVM discriminative model with a generative model (the HMM-based Fischer kernel). These methods can be quite effective for protein classification at the family level but they are often expensive to train and require some manual intervention to obtain optimal results (e.g., special input pre-processing). Other studies used unsupervised learning techniques, such as clustering (Yona et al., 1999; Tatusov et al., 1997) and self-organizing maps (Agrafiotis, 1997). While these methods can be applied successfully to broader data sets, they are not as accurate as the supervised-learning algorithms mentioned above.

Of all models, the HMM model is perhaps the most popular and successful model for protein classification to-date (Park et al., 1998). The specific class of HMMs that are used to model protein families (bio-HMM) has a special architecture that resembles the one used in speech-recognition. These models are first-order Markov models. The methods tested in this paper belong to the general class of HMMs. However, unlike the first-order bio-HMMs, the family of models tested here is capable of modeling higher order models. Moreover, in many cases these models obtain comparable performance to bio-HMMs (see the work of Bejerano & Yona, 2001) and are easier to train than bio-HMMs, which makes them an appealing alternative.

It should be noted that the classification task that we attempted here has a different setting than the typical setting in other studies on protein classification. In most studies, the classification focuses on the family level and a model is trained for each family. Here we focused on the structural fold-family level. This is a more difficult task and it has been the subject of strong interest in fold prediction tasks (CASP, 2002). Predicting the structural fold of a protein sequence is the first substantial step toward predicting the three-dimensional structure of a protein, which is considered one of the most difficult problems in computational biology.

The success rate of the VMM algorithms, and specifically the LZ-MS algorithm, is especially interesting since the sequences that make up a fold-family belong to different protein families with different properties. These families manifest different variations over the same structural fold; however, detecting these similarities has been proved to be a difficult problem (Yona & Levitt, 2002), and existing methods that are based on pure sequence information achieve only moderate success in this task.

Although we have not tested bio-HMMs, we suspect that such a model would be only moderately successful, since a single model architecture is incapable of describing the diverge set of protein families that make up a fold-family. In contrast, the VMM models tested here are capable of transcending the boundaries between different protein families since they do not limit themselves to a specific architecture. These models have very few hyper-parameters and they can be quite easily trained from the data without further pre-processing. Other studies that addressed a somewhat similar problem to ours used sequence comparison algorithms (McGuffin et al., 2001) or SVMs and Neural Networks (Ding & Dubchak, 2001) reporting a maximal accuracy of 56%. It is hard to compare our results to the results reported in these studies since they used different data sets and a different assessment methodology. However, the remarkably high success rate of the LZ-MS model in this task suggests that this model can be very effective for structural-fold prediction.

## 8. Concluding Remarks and Open Questions

In this paper we studied the empirical performance of a number of prominent prediction algorithms. We focused on prediction settings that are more closely related to those required by machine learning practitioners dealing with discrete sequences. Previous results related to these algorithms usually focused on a standard compression setting.

While it should not be expected that one algorithm will consistently outperform others on all tasks, our rather extensive empirical evaluations over the three domains indicate that there are prominent algorithms, which consistently tend to generate more accurate predictions than the other algorithms we examine. These algorithms are the 'prediction by partial match' (PPM-C) and 'decomposed context tree weighting' (DE-CTW). For classification of proteins we observe that there is also a consistent winner. However, somewhat surprisingly, the best predictor under the log-loss is not the best classifier. On the contrary, the consistently best protein classifier is based on the mediocre LZ-MS predictor! This algorithm is a simple modification of the well-known Lempel-Ziv-78 (LZ78) prediction algorithm, which can capture VMMs with large contexts. The surprisingly good classification accuracy achieved by this algorithm may be of independent interest to protein analysis research and clearly deserves further investigation.

This relative success of LZ-MS is related to the winner-takes-all classification scheme we use. A classifier based on this approach can suffer from a degraded performance if only one or a few of the models generated (for some of the classes) are wrongly fitted to other classes (in which case the erroneous model is the winner). Clearly, the superior performance of LZ-MS is related to its robustness in this regard and a closer inspection of the behavior of the algorithms may be revealing. A possible explanation of this robustness could be related to the unbounded memory length of the LZ-MS model. In most of the other VMM models the memory length is a hyper-parameter. However, as discussed in Section 7, some of the other algorithms such as PPM can be extended to work without a known bound on the context length.

An important observation, seen in these results, is that one cannot rely on log-loss (compression) performance when selecting a VMM algorithm for classification (using a WTA approach). Since the classification of sequences has not been studied as extensively as compression and prediction, we believe that much can be gained by focusing on specialized VMM algorithms for classification.

We hope that our contribution will assist practitioners in selecting suitable prediction algorithms for prediction and classification tasks and we also provide the code of all the algorithms we considered. Finally, we conclude by a number of open questions and directions for future research.

One of the most successful general-purpose predictors we examined is the DE-CTW algorithm, with its alphabet decomposition mechanism, as suggested by Volf (2002). As it turns out, this mechanism is crucial for the success of the the CTW scheme. However, currently there are no redundancy bounds (or other type of performance guarantees) for this decomposition. It would be interesting to prove such a bound, which will also motivate an optimality criterion for the decomposition-tree (see Section 3.4). Note that the heuristic proposed by Volf for generating the decomposition-tree relies on Huffman's coding, but we are not familiar with a compelling reason for the success of this approach.

Similarly, the PPM scheme (and the PPM-C variant we examined) do not have any known bounds on its log-loss redundancy. Clearly, such bounds could significantly contribute to understanding this algorithm (and its variants) and may help in designing stronger versions of this scheme.

One of the main factors that influence the success of the PPM algorithm is the details of its escape mechanism, including the allocation of probability mass for the 'escape' event. This probability allocation is, in essence, a solution to the zero frequency problem (also called "missing mass" problem). So far, there are no formal justifications for the performance of the more successful escape implementations. It would be interesting to see if some of the recent results on the missing mass problem, such as those presented by Orlitsky et al. (2003) and McAllester and Ortiz (2003), can help in devising a formal optimality measure for the PPM escape implementations.

We can view the CTW algorithm as an ensemble of numerous VMMs. However, the weights of the various VMM models are fixed. Various ensemble methods in machine learning such as boosting and online expert advice algorithms attempt to optimize the weights of the various ensemble members. Could we achieve better performance guarantees and better empirical performance using adaptive weight learning approaches? Volf (2002) has some results (on combining LZ78 and DE-CTW) indicating that this is a promising direction. Other possibilities would be to employ portfolio selection algorithms to dynamically change the weights of several models, along the lines suggested by Kalai et al. (1999).

## 9. Acknowledgments

We are very grateful to Gil Bejerano, Charles Bloom and Paul Volf who provided valuable advice and assistance. We would also like to thank Bob Carpenter for his PPMC compression java implementation, Moti Nisenson for his LZms java implementation and Adiel Ben Shalom for his MIDI assistance. Finally, we thank the anonymous referees for their good comments. The work of Ran El-Yaniv was partially supported by the Technion V.P.R. fund for the promotion of sponsored research and the partial support of the PASCAL network of excellence.

## Appendix A. Algorithm Hyper-Parameters Selection

In our experimental protocol each algorithm optimizes its hyper-parameters during the training phase. The best found set of parameters is then used for training (see Section 4 for details). This appendix specifies the possible hyper-parameter values that were considered for each algorithm for both the prediction and classification experiments. These values were selected (based on commonsense) before the start of any experiment. We also provide here, for each algorithm, the average values that were actually selected.

We present these hyper-parameter values in two tables: Table 8 for the prediction setup, and Table 9, for the protein classification setup. We used two different hyper-parameter sets for these two setups due to memory complexity issues. Specifically, in the classification setup, as described in Section 6, the length of training sequences is significantly larger than the length of training sequences in the prediction experiment of Section 4. For example, in the protein classification, class 'C' consists of 4303 sequences from 78 different 'folds' and the *average* size of a sequence in class 'C' is 265. Therefore, the size of the training sequence for this class is $912,236$. In comparison, the *maximal* training sequence length in the text prediction experiment is $\frac{785396}{2} = 392,698$. Also, recall that in each classification experiment, the number of constructed models is the number of 'folds' (e.g., 78 in class 'C'), which also increases the memory overhead.

The requirement to accommodate long training sequences (in the classification experiment) directly influences the space used by most of the algorithms. The lengths of training sequences described above made it impossible for us to complete the protein classification experiment using the set of feasible parameters as described in Table 8. Therefore, for classification we reduced the set of possible hyper-parmeter and, as can be seen, each parameter set in Table 9 is a subset of the corresponding set in Table 8. Finally, note that this table refers to a slight variant of the standard PST, denoted by PST∗, which was used in the protein classification experiment (see Section 6).

In Tables 8 and 9 we also present the average (overall runs) of the hyper-parameter values selected by the cross-validated optimization. For example, in Table 8 it is interesting to see that different values were selected for the LZ-MS algorithm, for different datasets. For protein sequence predictions LZ-MS is optimized with values that make it almost identical to the LZ78 algorithm (i.e. $M = 0$ and $S = 0$), while on the music predictions these parameters are optimized with significantly larger values ($M = 6.12$ and $S = 16.11$). Observe that the average order selected for PPM-C on 'text' data is 9.25. When originally introduced, PPM algorithms were applied with a hyper-parameter $D = 4$ (constrained by the memory available on computers at the time). Since then, it has often been used with a somewhat larger value, typically 5 or 6 and it was observed that performance begins to deteriorate when $D$ is increased beyond this. For example, see Cleary and Teahan (1997, Figure 2) for a drawing of compression vs. $D$ for *book1*. We observe very similar behavior when considering the English text files alone (not shown in our table). For example, the value selected the hyper-parameter $D$ in most text files (including *book1*) is 5.

The PST hyper-parameters presented in Tables 8 and 9 corresponds to the PST algorithm description in Section 3.5. Specifically, the $P_{min}$ (alternatively, `hits`) hyper-parameter is the threshold used for filtering "suffix set" candidates, $\alpha$ and $\gamma$ define the first condition of the PST second stage, $r$ (alternatively, $N_{min}$) defines the second condition of PST's second stage and $D$ defines the maximal order of the PST "suffix set".

| Algorithm | Hyper-parameter | Possible Values | Selected Values (Average) | | |
|---|---|---|---|---|---|
| | | | Text | Music | Protein |
| PPM-C | $D$ | $\{1, 3, \ldots, 19\}$ | 9.25 | 16.12 | 1.85 |
| BI-CTW | $D$ | $\{8, 16, 32, 64\}$ | 42 | 57.76 | 34.3 |
| DE-CTW | $D$ | $\{2, 4, 8, 16, 32\}$ | 7.06 | 27.97 | 17.9 |
| PST | $P_{min}$ | $\{0.0001, 0.001, 0.01, 0.1\}$ | 0.006 | 0.0003 | 0.001 |
| | $\alpha$ | $\{0\}$ | 0 | 0 | 0 |
| | $\gamma$ | $\{0.0001, 0.001, 0.01, 0.1\}$ | 0.0006 | 0.0005 | 0.0001 |
| | $r$ | $\{1.05\}$ | 1.05 | 1.05 | 1.05 |
| | $D$ | $\{5, 10, 15, 20\}$ | 7.33 | 15.6 | 20 |
| LZ-MS | $M$ | $\{0, 2, \ldots, 6, 8\}$ | 2 | 6.12 | 1.06 |
| | $S$ | $\{0, 2, \ldots, 16, 18\}$ | 8.4 | 16.11 | 0.29 |

Table 8: Prediction setup hyper-parameter values.

| Algorithm | Hyper-parameter | Possible Values | Selected Values (Average) |
|-----------|-----------------|-----------------|---------------------------|
| PPM-C | $D$ | $\{1, 3, 5, 7, 9\}$ | 8.5 |
| BI-CTW | $D$ | $\{8, 16, 32\}$ | 31 |
| DE-CTW | $D$ | $\{4, 8\}$ | 7.5 |
| PST* | hits | $\{2, 3, 4\}$ | 5 |
| | $\alpha$ | $\{0\}$ | 0 |
| | $N_{min}$ | $\{2, 3, 4, 5\}$ | 4 |
| | $\gamma$ | $\{0.001\}$ | 0.001 |
| | $r$ | $\{1.05\}$ | 1.05 |
| | $D$ | $\{10, 15, 20\}$ | 20 |
| LZ-MS | $M$ | $\{0, 2, 4, 6, 8\}$ | 5.9 |
| | $S$ | $\{0, 2, 4, 6, 8\}$ | 7.9 |

Table 9: Protein classification setup hyper-parameter values. Recall that in this setup we used a slight variation of the PST algorithm. The PST* variation uses the $N_{min}$ parameter that defines the threshold for a context candidate context; and the *hits* parameter defines the threshold for a context to be added to the PST tree.

## Appendix B. Music Representation

Our music data consists of polyphonic music pieces given as MIDI files. A polyphonic music piece is usually represented in a MIDI file as a collection of *channels*. Each channel is typically associated with a single musical instrument and includes instructions (called "events") regarding which notes to play and how long and loud to play them. We treat each channel as an individual sequence. For example, if a piece is played by five instruments, encoded in five MIDI channels, we generated five individual sequences, one for each channel. Note that each entry in Table 4 is an average of all channels of the corresponding piece.

We now describe how we represent each MIDI channel. Each note in a channel is represented as a triple: *pitch:volume:duration*. Both pitch and volume have, according to the MIDI protocol, 128 possible values. Duration is measured in milliseconds. We also include a special note, which we call a 'silence note' and allow for negative durations of this silence note. Such negative assignments are in fact negative time intervals, which make it possible to represent chords or other simultaneous melodic lines within the same channel.

The sequences corresponding to a channel are the ascii representation of the sequence of *pitch:volume:duration*. For example, the note 102:83:4022 is represented as "102:83:4022:".[29] Clearly, this representation is over an alphabet $\Sigma = \{0, 1, \ldots, 9, :, -\}$, of size is 12.

It is important to note that our representation preserves much of the information encoded in the original MIDI representation and, in reality, allows one to almost exactly reconstruct the original tunes.

## Appendix C. On the LZ-MS Hyper-Parameters

This appendix provides an indication on the effectiveness of the $M$ and $S$ hyper-parameters of the LZ-MS algorithm. The experimental setup is identical to the prediction setup described in Section 4. Table 10 shows the log-loss results over the Calgary Corpus of applications of the LZ-MS algorithm: in the second column only the $M$ parameter is enabled (and $S = 0$); and the third column provides the results when only the $S$ parameter is enabled (and $M = 0$). The losses in the last column correspond to applications where both $M$ and $S$ are enabled (this column is identical to the LZ-MS results in Table 3). It is evident that the best choice of $M$ alone is consistently better than the best choice of $S$ alone. However, the best combination of both $M$ and $S$ is consistently better than any of these parameters alone.

---

29. The first ten notes of Chopin's Etude 12 Op. 10 are: "83:120:240: 128:0:-240: 79:120:240: 128:0:-240: 77:120:240: 128:0:-240: 74:120:240: 128:0:-240: 71:120:240: 128:0:-180:".

| Sequence (length $\cdot 10^3$) | M (param. value) | S (param. value) | LZ-MS |
|---|---|---|---|
| $bib(111)$ | 2.67 (2) | **2.63** (18) | **2.26\*** (2,12) |
| $book1(785)$ | **2.76** (2) | 3.03 (18) | **2.57\*** (2,12) |
| $book2(611)$ | **2.80** (2) | 3.06 (18) | **2.63\***(2,12) |
| $geo(102)$ | 4.48 (0) | 4.48 (0) | 4.48 (0,0) |
| $news(377)$ | **3.19** (2) | 3.39 (12) | **3.07\*** (2,4) |
| $obj1(22)$ | **5.75\*** (2) | 6.24 (10) | **6** (2,2) |
| $obj2(247)$ | **3.54** (2) | 3.74 (18) | **3.48\*** (2,4) |
| $paper1(53)$ | **3.54** (2) | 3.72 (14) | **3.38\*** (2,12) |
| $paper2(82)$ | **3.10** (2) | 3.23 (18) | **2.81\*** (2,12) |
| $paper3(47)$ | **3.32** (2) | 3.44 (18) | **3.09\*** (2,12) |
| $paper4(13)$ | **3.68** (2) | 3.79 (10) | **3.61\*** (2,14) |
| $paper5(12)$ | **4.26** (2) | 4.42 (18) | **4.16\*** (2,8) |
| $paper6(38)$ | **3.70** (2) | 3.91 (10) | **3.6\*** (2,12) |
| $pic(513)$ | 0.79 (2) | 0.79 (4) | 0.79 (2,0) |
| $progc(40)$ | **3.34** (2) | 3.57 (10) | **3.22\*** (2,4) |
| $progl(72)$ | **3.26** (2) | 3.50 (18) | **3.26** (2,16) |
| $progp(49)$ | **2.75** (2) | 2.93 (18) | **2.65\*** (2,18) |
| $trans(94)$ | **2.85** (2) | 2.89 (16) | **2.61\*** (4,18) |
| Average | **3.32 ± 0.06** | 3.48 ± 0.06 | **3.2 ± 0.06\*** |

Table 10: Comparing effectiveness of the LZ-MS hyper-parameters. Second column: $M$ is assigned the best value from the "feasible" set $\{0, 2, 4, 6, 8\}$ and $S = 0$. Third column: $S$ is assigned the best value from $\{0, 2, \ldots, 16, 18\}$ and $M = 0$. Last column: the best combination of both $M$ and $S$ is selected from the two feasible sets, respectively. Best values are determined using a five-fold CV. Each entry in this table provides the average log-loss and the corresponding value of the hyper-parameter that was used for obtaining this log-loss. For example, the 2.63 loss for the bib file when optimizing $S$ alone is obtained with $S = 18$.